%% file: main.tex
\documentclass[10pt,twocolumn,letterpaper]{article}

\usepackage[pagenumbers]{cvpr} 

\usepackage{graphicx}
\usepackage{amsmath}
\usepackage{amssymb}
\usepackage{booktabs}
\usepackage{anyfontsize}

\usepackage[pagebackref,breaklinks,colorlinks]{hyperref}

\usepackage[capitalize]{cleveref}
\usepackage{import}
\usepackage{arydshln}
\usepackage{multirow}
\usepackage{lipsum}
\usepackage{inconsolata} 
\usepackage{amssymb}
\usepackage{pifont}
\usepackage{clock}
\usepackage[clock]{ifsym}
\usepackage{float}
\usepackage{graphicx}
\usepackage{amsmath}
\usepackage{soul}
\usepackage[table]{xcolor}
\usepackage{enumerate}
\usepackage{enumitem}
\usepackage{subcaption}
\usepackage{adjustbox}

\def\cvprPaperID{6908} 
\def\confName{CVPR}
\def\confYear{2023}

\crefname{section}{Sec.}{Secs.}
\Crefname{section}{Section}{Sections}
\Crefname{table}{Table}{Tables}
\crefname{table}{Tab.}{Tabs.}

\definecolor{darkgreen}{rgb}{0.0, 0.75, 0.0}

\definecolor{bananayellow}{rgb}{1.0, 0.88, 0.21}

\definecolor{green}{RGB}{0,195,0}
\definecolor{red}{RGB}{195,0,0}
\definecolor{mygray}{RGB}{232, 234, 237}
\definecolor{lightorange}{RGB}{252,228,181}
\definecolor{lightskyblue}{RGB}{210,238,247}
\definecolor{myblue}{HTML}{006EB8}
\newcommand{\graymidrule}{
    \arrayrulecolor{gray}\midrule\arrayrulecolor{black}
}
\newcommand{\lgraymidrule}{
    \arrayrulecolor{lightgray}\midrule\arrayrulecolor{black}
}
\newcommand{\lgraycmidrule}[1]{
    \arrayrulecolor{lightgray}\cmidrule{#1}\arrayrulecolor{black}
}
\newcommand{\cmark}{\ding{51}}%
\newcommand{\xmark}{\ding{55}}%

\newlength\myheight%
\newlength\mydepth%
\settototalheight\myheight{Xygp}
\newcommand{\marktext}[2]{\adjustbox{bgcolor=#1}{\strut\ #2}}

\colorlet{shadecolor}{gray!25}
\renewcommand{\paragraph}[1]{\vspace{1.5mm}\noindent\textbf{#1}}

\newcommand{\bz}{\mathbf{z}}
\newcommand{\mcL}{\mathcal{L}}
\newcommand{\mcB}{\mathcal{B}}
\newcommand{\mcC}{\mathcal{C}}
\newcommand{\mcV}{\mathcal{V}}
\newcommand{\mcD}{\mathcal{D}}
\newcommand{\tnce}{\operatorname{TNCE}}
\newcommand{\trev}{\mathbb{T}}
\newcommand{\Atime}{A_{\text{time}}}
\newcommand{\Ctime}{\mcC^{\text{time}}}
\newcommand{\Deltatime}{\Delta_\text{time}}
\newcommand{\alphasame}{\alpha_{\text{same}}}
\newcommand{\alphacross}{\alpha_{\text{cross}}}
\newcommand{\clipartPolice}{
    \raisebox{-\mydepth}{\fbox{\includegraphics[height=3mm]{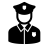}}}
}
\newcommand{\clipartClock}{
    \raisebox{-\mydepth}{\fbox{\includegraphics[height=3mm]{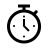}}}
}

\begin{document}

\title{Test of Time: Instilling Video-Language Models with a Sense of Time}

\author{
    Piyush Bagad\\
    \normalsize University of Amsterdam\\
    \and
    Makarand Tapaswi\\
    \normalsize IIIT Hyderabad\\
    {
        \tt\small%
        \href{https://bpiyush.github.io/testoftime-website/}{
            bpiyush.github.io/testoftime-website
        }
    }
    \and
    Cees G.M. Snoek\\
    \normalsize University of Amsterdam\\
}
\maketitle

\import{sections}{abstract}

\import{sections}{intro}
\import{sections}{related_work}
\import{sections}{motivation}
\import{sections}{method}
\import{sections}{expts-adaptation}
\import{sections}{expts-downstream}
\import{sections}{discussion}


{\small
\bibliographystyle{ieee_fullname}
\bibliography{longstrings,egbib}
}

\import{sections}{supp}

\end{document}

%% file: sections/abstract.tex
\begin{abstract}
Modelling and understanding time remains a challenge in contemporary video
understanding models. With language emerging as a key driver towards powerful
generalization, it is imperative for foundational video-language models to have
a sense of time. In this paper, we consider a specific aspect of temporal
understanding: consistency of time order as elicited by before/after relations.
We establish that seven existing video-language models struggle to understand
even such simple temporal relations. We then question whether it is feasible to
equip these foundational models with temporal awareness without re-training them
from scratch. Towards this, we propose a temporal adaptation recipe on top of
one such model, VideoCLIP, based on post-pretraining on a small amount of
video-text data. We conduct a zero-shot evaluation of the adapted models on six
datasets for three downstream tasks which require varying degrees of time
awareness. We observe encouraging performance gains especially when the task
needs higher time awareness. Our work serves as a first step towards probing and
instilling a sense of time in existing video-language models without the need
for data and compute-intense training from scratch.
\end{abstract}

%% file: sections/intro.tex
\vspace{-0.7em} 
\section{Introduction}
\vspace{-0.2em}
Self-supervised pretraining at scale on multimodal web corpora tied with
powerful architectures~\cite{transformers-NIPS2017_3f5ee243} has led to
foundational models~\cite{Bommasani2021OnTO-foundation} for
images~\cite{il1-Radford2021LearningTV,il2-Jia2021ScalingUV,
il4-Alayrac2022FlamingoAV,dalle-Ramesh2021ZeroShotTG,Li2022BLIPBL}
and
videos~\cite{il4-Alayrac2022FlamingoAV,xu-etal-2021-videoclip,Wang2022OmniVLOF,
Bain21-frozen,Yuan2021FlorenceAN,Fu2021VIOLETE}.
These models have enabled remarkable improvements on a plethora of downstream
video-language tasks such as video-text retrieval, video question-answering, and
action recognition. Given the cost and difficulty of video annotations, even for
a small amount of downstream data, such foundational models are emerging as the
de-facto backbone for
zero-shot~\cite{xu-etal-2021-videoclip,yang2022frozenbilm,zeng2022socraticmodels}
and few-shot generalization~\cite{il4-Alayrac2022FlamingoAV}. However, it
remains unclear if these video-language models capture essential properties of a
video beyond what can be learned from static images, most notably:
\textit{time}.

Many before us have shown that existing video-language
models~\cite{xu-etal-2021-videoclip,Bain21-frozen,Luo2021-CLIP4Clip,clipbert-Lei2021LessIM}
can achieve impressive performance on several video
benchmarks~\cite{xu2016-msr-vtt,caba2015-activitynet,didemo-Hendricks2017LocalizingMI}
without reliably encoding
time~\cite{buch2022-revisiting,revealing-single-frame,Li2022BLIPBL}. For
example, Buch \etal ~\cite{buch2022-revisiting} show that a model that uses a
single (carefully selected) frame  often outperforms recent video-language
models~\cite{clipbert-Lei2021LessIM,xu-etal-2021-videoclip} on standard video
benchmarks such as MSR-VTT~\cite{xu2016-msr-vtt}. Lei \etal
~\cite{revealing-single-frame} report similar findings with a single-frame
pretraining approach. These findings hint at a lack of time awareness in video
models. However, it remains unclear if these findings are caused, indeed, by the
lack of time in video models or whether the benchmarks themselves do not mandate
time awareness. Furthermore, there is no clear definition of what it means for a
model to be time aware. In this paper, we strive to shed light on all these
factors of time awareness in video-language models.

\begin{figure}[t!]
\captionsetup{skip=1mm,font=small}
    \centering
    \includegraphics[width=\columnwidth]{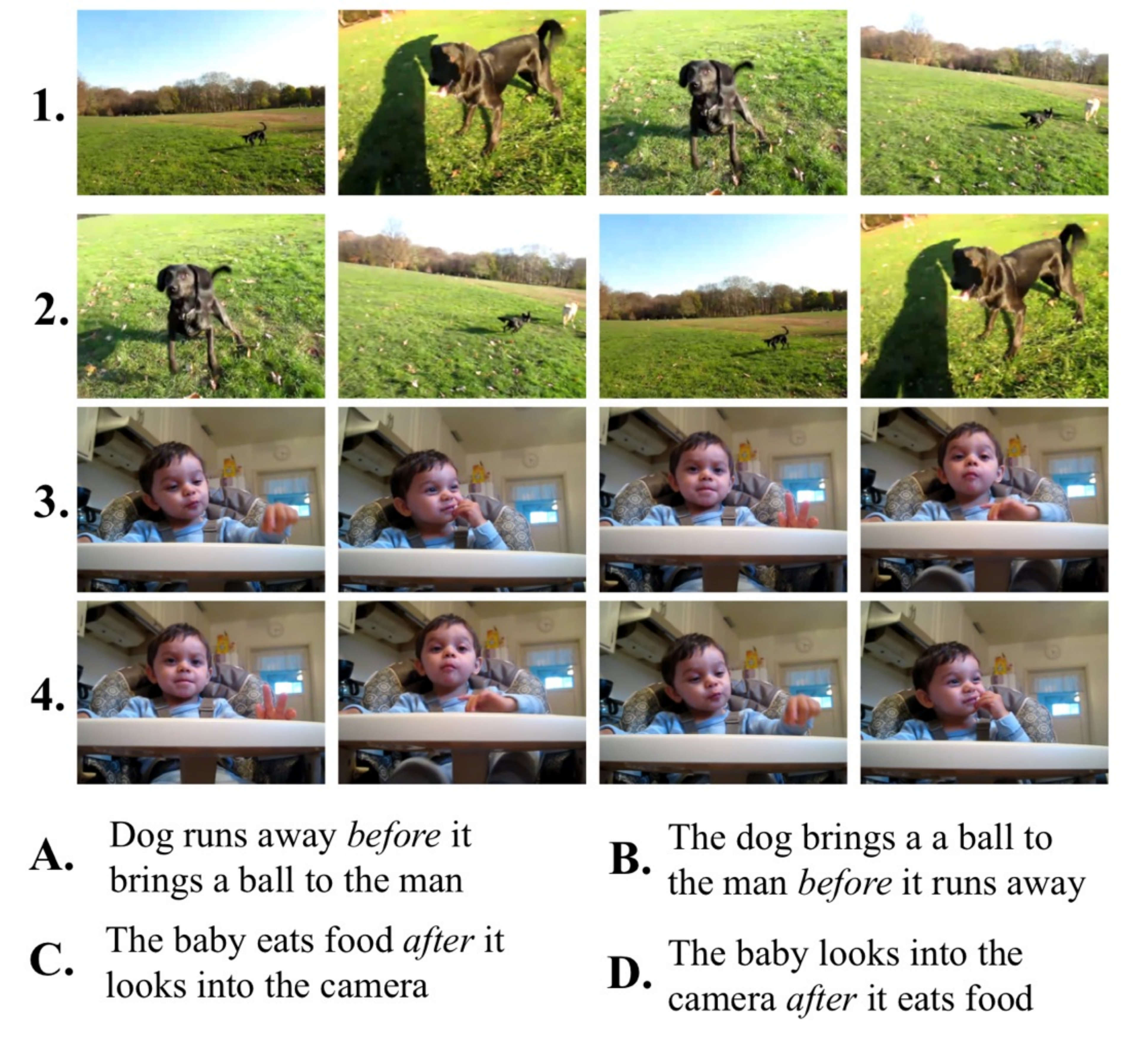}
    \caption{
        Can you match the correct video-text pairs?
        Understanding the time order of events across video and language
        is necessary to be able to solve this task. See footnote on next page for answers.
    }
    \label{fig:teaser}
\end{figure}

As a first step, we consider a simple notion of understanding time, \ie,
understanding temporal relations such as \textit{before} and \textit{after}
\cite{time-in-lang2-Allen1984TowardsAG}. Consider the task presented in
\cref{fig:teaser}. A time invariant model shall be able to associate (A) with
(1) or (2) and (B) with (3) or (4) based on static frames alone. But to
distinguish between (1) and (2), one needs to be able to understand time order
and connect it across video and language\footnote{Answers: (A)-(2), (B)-(1),
(C)-(4), (D)-(3).}. Thus, the \textbf{\textit{first question}} we ask in
Section~\ref{sec:motivation}: do the representations learnt by foundational
video-language models encode this sense of time? To reliably attribute lack of
time awareness to models and not existing benchmarks, we design our own
synthetic dataset to probe models for this sense of time. We create
video-language pairs that show a sequence of two events. Then, we alter the
order of events either in the text or the video and check if models can connect
the order in video and language. We find that existing video-language models
indeed struggle to associate the time order across video and
language.

In light of these findings, the \textbf{\textit{second question}} we ask in
Section~\ref{sec:tact} is: can we adapt a video-language model, without
expensive re-training from scratch, to instill this sense of time? Towards this,
we take inspiration from literature on understanding time in natural language,
where there has been much work on developing time aware language
models~\cite{tlb1-Dhingra2022TimeAwareLM,tr2-Han2021ECONETEC,tr1-Han2020DEERAD,
tr4-Zhou2020TemporalCS,tr3-Zhou2021TemporalRO}.
Our objective is to instill time awareness in a video-language model without
having to pretrain from scratch. To do that, we propose TACT: \textbf{T}emporal
\textbf{A}daptation by \textbf{C}onsistent \textbf{T}ime-ordering based on two
key components: (i)~we artificially create samples that provide temporal signal,
for example, by flipping the order of events in the video or the text, and
(ii)~we introduce a modified contrastive loss to learn time order consistency
based on these samples. Instead of training from scratch, we adapt an existing
video-language model,
VideoCLIP~\cite{video-continuity1-Liang2022SelfsupervisedSR}, using the paradigm
of \textit{post-pretraining} on a \textit{small} amount of video-text
data~\cite{Luo2021-CLIP4Clip, ilv5-Xue2022CLIPViPAP}. We demonstrate the
effectiveness of TACT in connecting the time order in video and language on four
diverse real datasets in Section~\ref{sec:ablations}.

Finally, in line with the original motive of video-language models for zero-shot
generalization, we evaluate in Section~\ref{sec:downstream} our TACT-adapted
model for three sets of tasks on six downstream datasets which require a varying
degree of time awareness. On tasks that need higher time awareness, with the
appropriate choice of adaptation dataset, TACT outperforms a strong baseline
that is based on post-pretraining on canonical clip-text pairs without
consideration of time-order. the corresponding results, we first provide a
broader background on related work.

In summary, our contributions are: (i)~We show that existing video-language
models struggle to associate time order in video and language through controlled
experiments on synthetic data and several evaluations on real datasets.
(ii)~Based on VideoCLIP~\cite{xu-etal-2021-videoclip}, we propose TACT, a method
for temporal adaptation using this time order consistency without having to
pretrain from scratch. (iii)~We demonstrate improved zero-shot generalizability
of TACT-adapted models on tasks that require higher time awareness.

%% file: sections/related_work.tex
\section{Background and Related Work}
\label{sec:related_work}

We briefly discuss recent advances in video-language models followed by their consideration of time.

\paragraph{Foundational video-language models.}
Large-scale datasets, self-supervision, and the advent of
Transformers~\cite{transformers-NIPS2017_3f5ee243} have led to the emergence of
powerful encoders for
images~\cite{ie1-he2016residual,ie2-dosovitskiy2020vit,ie3-tolstikhin2021mixer},
videos~\cite{ve1-Tran2018ACL,ve2-Xie2017RethinkingSF,ve3-Feichtenhofer2019SlowFastNF,
ve4-Bertasius2021IsSA,ve5-Arnab2021ViViTAV},
language
models~\cite{te1-Mikolov2013EfficientEO,te2-devlin-etal-2019-bert,
te3-Liu2019RoBERTaAR,te4-Sanh2019DistilBERTAD}
and even universal encoders
~\cite{ue1-Jaegle2021PerceiverGP,ue2-Girdhar2022OmnivoreAS}. These encoders form
the basis for several vision-language foundational models. Popular
image-language models such as CLIP~\cite{il1-Radford2021LearningTV} and
ALIGN~\cite{il2-Jia2021ScalingUV} are trained on massive datasets by using web
images and alt-text. Similarly, video-language models are catching up and can be
categorised into two broad directions: (i)~adapting image-language models for
videos~\cite{Luo2021-CLIP4Clip,fang2021-clip2video,clip-H,ilv1-Lin2022FrozenCM,
ilv2-XCLIP,ilv3-Ju2021PromptingVM,ilv4-Wang2022OmniVLOF,ilv5-Xue2022CLIPViPAP,
ilv6-Wang2022BEVTBP,Ju2021PromptingVM},
and (ii)~pure video-based models that are learned using large video-text
datasets~\cite{vlm0-Miech2020EndtoEndLO,miech19-howto100m,webvid-Bain21,
xu-etal-2021-videoclip,Ge_2022_CVPR-bridgeformer,vlm1-Sun2019VideoBERTAJ,
vlm2-Ging2020COOTCH,vlm3-Gabeur2020MultimodalTF,vlm4-Alayrac2020SelfSupervisedMV,
vlm5-Luo2020UniVL,vlm6-Li2020UnicoderVLAU,vlm7-Fu2021VIOLETE,vlm8-Lin2022EgocentricVP}.
Recently, a new paradigm of \textit{post-pretraining} has emerged where an
existing image- or video-language model goes through another stage of
self-supervised pretraining on a \textit{small} amount of video data before it
is evaluated on downstream tasks~\cite{Luo2021-CLIP4Clip,ilv5-Xue2022CLIPViPAP}.
This is promising as it circumvents the prohibitive cost of pretraining on large
datasets from scratch. In~\cite{Luo2021-CLIP4Clip}, the post-pretraining uses
time-invariant mean-pooling, while \cite{ilv5-Xue2022CLIPViPAP} strives to
bridge the domain gap between image captions and video subtitles. In contrast,
our proposed temporal adaptation involves post-pretraining of
VideoCLIP~\cite{xu-etal-2021-videoclip} with a small amount of data that
instills the model to learn the time-order of events in a video. 

\paragraph{Time in vision.}
Time separates videos from static images or an unordered set of frames. While
modeling time remains a challenge, it also presents a natural source of
supervision that has been exploited for self-supervised learning. For example,
as a proxy signal by posing pretext tasks involving spatio-temporal
jigsaw~\cite{jigsaw1-ahsan2019video, jigsaw2-huo2021selfsupervised,
jigsaw3-kim2019self}, video
speed~\cite{relative-speed1-benaim2020speednet,relative-speed2-jenni2020video,
relative-speed3-Singh2021SemiSupervisedAR,playback1-cho2020self,
playback2-yao2020video, playback3-wang2020self}, arrow of
time~\cite{aot1-wei2018learning,aot2,aot3-Price2019RetroActionsL}, frame/clip
ordering~\cite{frame-order-misra2016shuffle, shuffle1-fernando2017self,
shuffle2-suzuki2018learning, clip-order-xu2019self, sharma2019video-ordering},
video continuity~\cite{video-continuity1-Liang2022SelfsupervisedSR}, or
tracking~\cite{ctp,tracking2-Jabri2020SpaceTimeCA,tracking3-Wang2019LearningCF}.
Several works have also used contrastive learning to obtain spatio-temporal
representations by (i)~contrasting temporally augmented versions of a
clip~\cite{clt1-Jenni2021TimeEquivariantCV,
clt2-Qian2021SpatiotemporalCV,clt3-Pan2021VideoMoCoCV},
or (ii)~encouraging consistency between local and global temporal
contexts~\cite{Recasens2021BroadenYV,clt4-Behrmann2021LongSV,
clt6-Dave2022TCLRTC,clt5-Yang2020BackTT}.
Nevertheless, it remains unclear if the learnt representations actually encode
time reliably. Time-aware features have also been explored for specific
downstream tasks such as action recognition~\cite{Ghodrati2018VideoTP,
Thoker2021SkeletonContrastive3A, thoker2023tubeletcontrastive}. There has been
some very recent work on evaluating self-supervised video
representations~\cite{survey1-ChantrySchiappa2022SelfSupervisedLF,
survey2-Thoker2022HowSI}
on their temporal recognition ability instead of only relying on time as a
guidance for training.

In the same spirit, a related direction pursues evaluation and benchmarking of
time awareness in video
datasets~\cite{DBLP:journals/corr/abs-1907-08340-temporal-dataset},
models~\cite{Ghodrati2018VideoTP,
revealing-single-frame,buch2022-revisiting,imgclf-time-Shankar2021DoIC,
Yi2020CLEVRERCE,BYVSHEV2022103437}
or both~\cite{8578867-what-makes-video,Sigurdsson2017What-actions}.
Huang~\etal~\cite{8578867-what-makes-video} measure the effect of motion on
temporal action recognition to find that only a subset of classes in UCF-101 and
Kinetics-400 require motion information.
Ghodrati~\etal~\cite{Ghodrati2018VideoTP} propose new tasks to evaluate temporal
asymmetry, continuity and causality in video models. Our work derives
inspiration from these but applies more generally to video-language models as
language provides a basis for open-world generalization.

\paragraph{Time in language.}
Time has also been extensively studied in the natural language literature. Early
works identified temporal structures in language such as temporal prepositions
and
quantifiers~\cite{time-in-lang1-Pratt2001TemporalPA,time-in-lang2-Allen1984TowardsAG}.
More recent literature focuses on tasks such as extracting temporal
relations~\cite{utr1-Ning2019AnIN,utr2-Ning2018JointRF,utr3-Ning2017ASL,utr4-Han2019JointEA},
as well as temporal
reasoning~\cite{tr1-Han2020DEERAD,tr2-Han2021ECONETEC,tr3-Zhou2021TemporalRO,
tr4-Zhou2020TemporalCS,tr5-Qin2021TIMEDIALTC}.
For example, Han~\etal~\cite{tr2-Han2021ECONETEC,tr1-Han2020DEERAD} and
Zhou~\etal~\cite{tr3-Zhou2021TemporalRO} pretrain language models specifically
to focus on understanding temporal relations such as before, after, during,
\etc. The emergence of large language models has also spurred an increased
interest in developing benchmarks to test for time awareness in these
models~\cite{tlb1-Dhingra2022TimeAwareLM,tlb2-Thukral2021ProbingLM,tlb3-Ning2020TORQUEAR,
tlb4-Zhou2019GoingOA,tlb5-Vashishtha2020TemporalRI,tlb6-Ning2018CogCompTimeAT}.
For example, Ning~\etal~\cite{tlb3-Ning2020TORQUEAR} propose a new benchmark of
reading comprehension with questions involving before/after relations. Since
temporal relations in language are grounded in the video, we draw inspiration
from~\cite{tr2-Han2021ECONETEC,tr1-Han2020DEERAD,tr3-Zhou2021TemporalRO} and aim
to instill time awareness in video-language models.

\paragraph{Time in video-language models}
appears implicitly in tasks like video-text
alignment~\cite{vl-align1-Han2022TemporalAN, Tapaswi_2015_CVPR} and temporal
grounding~\cite{time-ground1-Hendricks2018LocalizingMI,Li2022CompositionalTG}.
In this work, we focus on self-supervised video-language models that can
generalize to a variety of tasks rather than models designed for a specific
task, \eg, temporal grounding. Some recent works have shown the
under-utilisation of time in classic video-text benchmarks such as
MSR-VTT~\cite{xu2016-msr-vtt}, YouCook~\cite{ZhXuCoAAAI18-youcook2},
ActivityNet~\cite{caba2015-activitynet}, and
DiDeMo~\cite{didemo-Hendricks2017LocalizingMI}. For example,
\cite{buch2022-revisiting,revealing-single-frame,clipbert-Lei2021LessIM}~discover
that on several benchmarks, using only one or few frames or clips achieves
competitive performance. Adaptations of the popular CLIP architecture for videos
(\eg,~CLIP4Clip~\cite{Luo2021-CLIP4Clip}) show that weighted pooling of frames
already achieves impressive performance on retrieval benchmarks.

These raise some key questions: do existing video-language models truly
understand time  in the sense of correctly associating order of events in
language and video? If not, can we adapt them to instill time awareness? Our
work addresses these questions. There has been some work in using time-order
across video and language as a source of self-supervision. Specifically,
concurrent to our work, both Sun~\etal~\cite{temporal1-sun2022longform} and
Cao~\etal~\cite{temporal2-cao2022locvtp} propose fine-grained temporal alignment
between video and text as the pretraining objective. Different from these works,
we consider the notion of time-order and we aim to adapt a given video-language
model using \textit{post-pretraining}, which circumvents the need for a new
round of compute-intense pretraining.

%% file: sections/motivation.tex
\section{Do Video-Language Models Sense Time?}
\label{sec:motivation}

Probing video-language models for temporal understanding is an open direction of
research. In this work, we consider a specific sense of temporal understanding:
consistency in the order of events in a video with the associated textual
description. For example, consider a text description: \texttt{A red circle
appears before a yellow circle}. This imposes an order constraint on the video
stream to have the event \texttt{red circle appears} happen before the event
\texttt{yellow circle appears}. Can existing video-language models connect
time-order in text with that in video? To answer this, we design an experiment
with synthetic data.

\begin{figure}[t]
\captionsetup{skip=2mm,font=small}
\centering
\includegraphics[width=\linewidth]{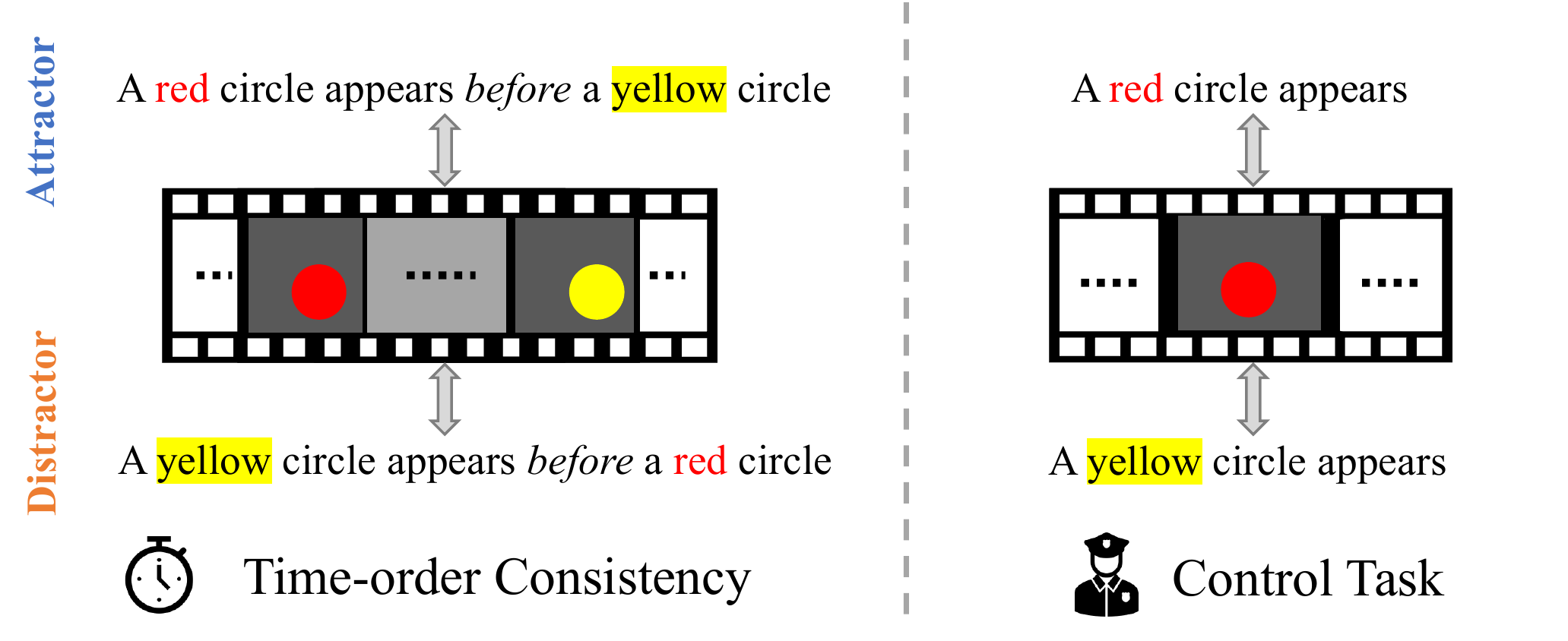}
\caption{Overview of the proposed task to evaluate time-order consistency across
synthetic video-language pairs having before/after relations. We also define a
control task to check if the synthetic videos are considered out-of-distribution
by the model. }
\label{fig:syn-task}
\end{figure}

\paragraph{Synthetic dataset.}
We construct simple videos that comprise of a pair of events such as the ones
mentioned above. We generate $N {=} 180$ video-language pairs as a combination
of $C {=} 6$ colors, $S {=} 3$ shapes, and $|\tau| {=} 2$ temporal relations:
\textit{before} and \textit{after}. The corresponding caption describes the
order of events connected with a \textit{before/after} temporal relation. We
call this caption as an \textit{attractor} since it is consistent with the
time-ordering in the video. Likewise, we construct a \textit{distractor} where
we flip the order of event descriptions while retaining the temporal relation.
An example pair is illustrated in \cref{fig:syn-task} (left). Ideally, a time
aware video-language model should be able to associate the video with the
temporally consistent text, or vice versa. We refer to this task as
\emph{time-order consistency check}. To rule out the possibility that synthetic
videos are out-of-distribution, we also perform the same experiment with
canonical clips with a single event and the text describes that same event as
shown in \cref{fig:syn-task} (right). We refer to this as the \emph{control
task}.

\paragraph{Choice of models.}
We consider recent video-language models, broadly categorized into three groups:
(i)~image-language models like CLIP~\cite{il1-Radford2021LearningTV} that are
adapted to videos~\cite{Luo2021-CLIP4Clip,fang2021-clip2video,2022-centerclip},
(ii)~pure video-language models trained on a contrastive learning
objective~\cite{xu-etal-2021-videoclip,Bain21-frozen,cheng2022vindlu}, and
(iii)~pure video-language models trained on a masking
objective~\cite{Ge_2022_CVPR-bridgeformer}.

\paragraph{Findings.}
We evaluate video-to-text and text-to-video retrieval on both time-order
consistency and control tasks. From \cref{tab:synthetic-v2t}, we observe that
while most video-language models perform well on the control task, all of them
struggle and perform on par with random chance on the temporal task. This gap in
performance deserves attention given the importance of time in videos. Note that
while synthetic data allows for controlled experiments, we also expand this
evaluation to real video datasets in the following section.

\input{tables/1_synthetic-expt}

%% file: tables/1_synthetic-expt.tex
\begin{table}[t!]
\captionsetup{skip=2mm,font=small}
\centering
\resizebox{\columnwidth}{!}{%
\begin{tabular}{llrrrr}
\toprule
\multicolumn{1}{l}{\textbf{Paradigm}} & \multicolumn{1}{l}{\textbf{Method}} & \multicolumn{2}{c}{\textbf{Video-to-Text}} & \multicolumn{2}{c}{\textbf{Text-to-Video}} \\
\midrule
\multicolumn{1}{c}{\textbf{}}
&
\multicolumn{1}{c}{\textbf{}}
&
\multicolumn{1}{r}{\includegraphics[height=4mm]{media/police.png}}
&
\multicolumn{1}{r}{
\includegraphics[height=4mm]{media/clock.png}
}
&
\multicolumn{1}{r}{\includegraphics[height=4mm]{media/police.png}}
&
\multicolumn{1}{r}{
\includegraphics[height=4mm]{media/clock.png}
} \\
\midrule
Chance & - & 50.0 & 50.0 & 50.0 & 50.0 \\
\arrayrulecolor{lightgray}\midrule
\multirow{3}{*}{\begin{tabular}[c]{@{}l@{}}Image-Language\\ adapted to video\end{tabular}} & CLIP4Clip~\cite{Luo2021-CLIP4Clip} & 49.4 & 51.1  & 50.0 & 49.4 \\
 & CLIP2Video~\cite{fang2021-clip2video} &  100.0 & 47.8  & 97.8 & 52.3 \\
 & CenterCLIP~\cite{2022-centerclip} &  91.7 & 46.1  & 97.2 & 51.1 \\
\arrayrulecolor{lightgray}\midrule
\multirow{2}{*}{\begin{tabular}[c]{@{}l@{}}Video-Language\\ Contrastive\end{tabular}} & VideoCLIP~\cite{xu-etal-2021-videoclip} &  87.1 & 51.1  & 66.7 & 48.3 \\
 & Frozen in Time~\cite{Bain21-frozen} &  97.8 & 49.4  & 100.0 & 50.6 \\
 & VindLU~\cite{cheng2022vindlu} &  98.0 & 52.0  & 100.0 & 51.1 \\
\arrayrulecolor{lightgray}\midrule
\begin{tabular}[c]{@{}l@{}}Video-Language\\ Masking\end{tabular} & BridgeFormer~\cite{Ge_2022_CVPR-bridgeformer} &  100.0 & 51.1  & 97.2 & 42.2 \\
\arrayrulecolor{black}\bottomrule
\end{tabular}%
}
\caption{
Results on synthetic control (\clipartPolice) and  time-order consistency (\clipartClock) task as described in \cref{fig:syn-task}.
Existing video-language models show random performance on our time-order task.
}
\label{tab:synthetic-v2t}
\end{table}

%% file: sections/method.tex
\section{Adaptation by Consistent Time-Ordering}%
\label{sec:tact}

We describe a post-pretraining recipe to instill a video-language model with a
sense of time. We propose TACT\: \textbf{T}emporal \textbf{A}daptation by
\textbf{C}onsistency of \textbf{T}ime-order, that is based on two key
components: (i)~we artificially create samples that provide temporal signals,
\eg,~by flipping the order of events; and (ii)~we introduce a modified
contrastive loss to learn temporal consistency based on these samples. We start
by defining the notation and then describe the key components of our adaptation
recipe.

\paragraph{Preliminaries.}
Let $\mathcal{V}$ be the space of video clips and $\mathcal{T}$ be the space of
text clips. Consider two non-overlapping video clips $v_{i}, v_{j} \in
\mathcal{V}$. Let $\zeta_{i}, \zeta_{j} \in \mathcal{T}$ be their respective
captions. Let $\tau$ be a temporal relation, $\tau \in \{\textit{before},
\textit{after}\}$. Then, we denote a \textit{stitched} and time-order consistent
clip as $(u_{ij}, t_{ij})$, where $u_{ij} := [v_{i};v_{j}]$, $t_{ij} :=
[\zeta_{i}; \tau; \zeta_{j}]$, and $[\cdot;\cdot]$ denotes concatenation. Note
that depending on $\tau$, the order of $v_i$ and $v_j$ may need to change in
$u_{ij}$. For brevity, we drop the subscripts and refer to the stitched pair as
$(u, t)$ unless stated otherwise.

\paragraph{Time-order reversal.}
The classic contrastive learning paradigm for video-language models aligns
components of a video clip $v_i$ with its text counterpart $\zeta_{i}$ and
contrasts against other texts $\zeta_{j}$ that usually describe a completely
different clip. This makes such models ignore the finer details of temporal
understanding as it is easier to contrast the negatives by simply focusing on
the objects or the scene. This is evident from simple bag-of-word-like methods
that are shown to work well for contrastive learning, both on the visual
(\eg,~CLIP4Clip~\cite{Luo2021-CLIP4Clip}) and textual
(\eg,~MIL-NCE~\cite{vlm0-Miech2020EndtoEndLO}) modalities. We hypothesize that
unless there are negatives in a contrastive setup that contain the same scenes
and objects, models need not learn a sense of time. Thus, we present a simple
strategy to generate negatives that force the learning process to focus on the
temporal order.

\begin{figure}[t!]
\captionsetup{skip=2mm,font=small}
\centering
\includegraphics[width=0.88\linewidth]{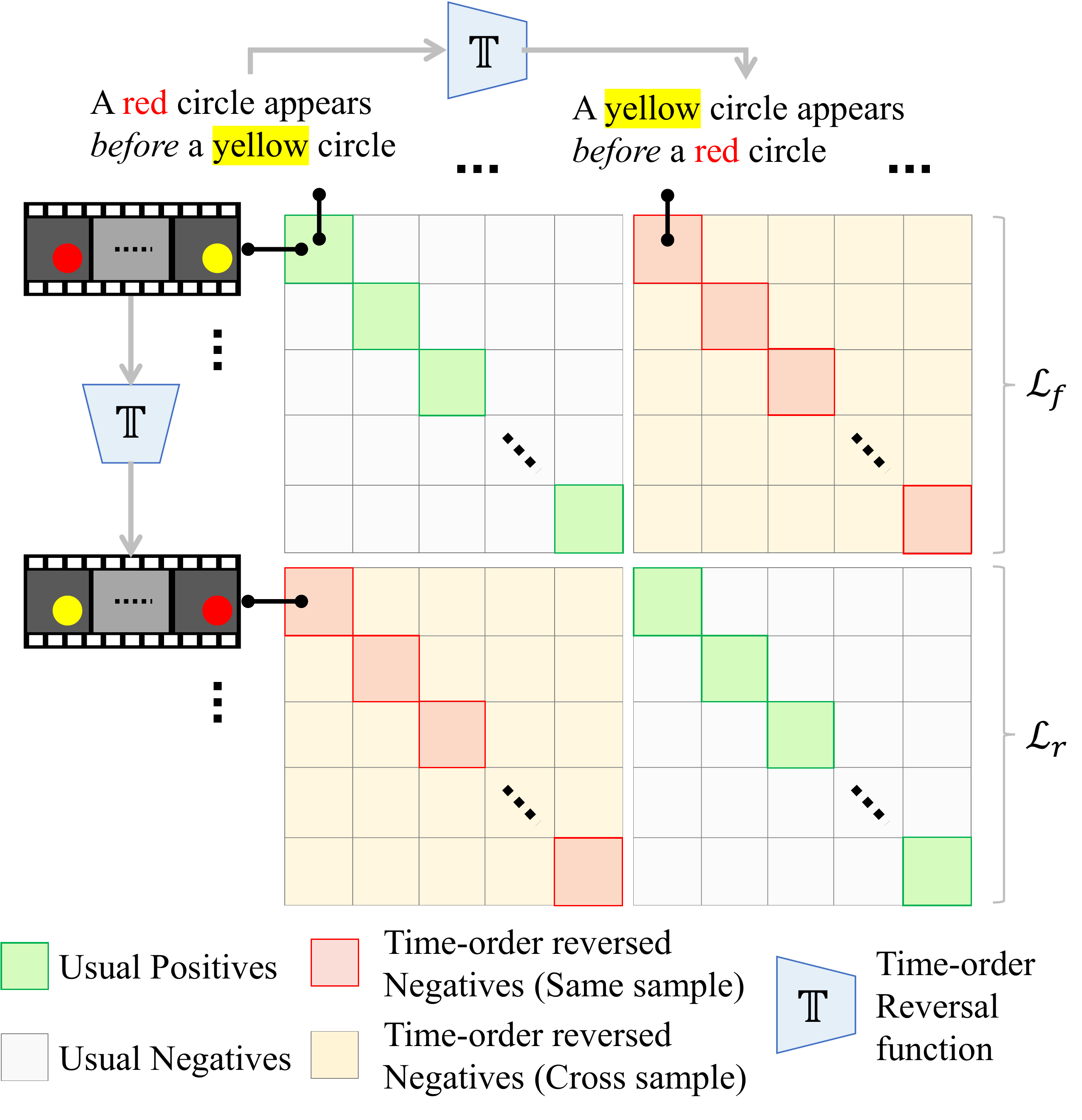}
\caption{Overview of TACT. Along with the usual contrastive loss, where
negatives come from other samples in the batch, we make use of time-order
reversal within the \colorbox{red!15}{same} sample and
\colorbox{lightorange}{cross} samples to generate additional negatives for both
video and text. We also extend the contrastive loss to time-order reversed video
and text corresponding to reverse consistency $\mcL_r$.
}%
\label{fig:loss_viz}
\end{figure}

We define a time-order reversal function $\trev$ that operates on the stitched
video clip or text description and temporally swaps its constituents:
\vspace{-1mm}
\begin{align}
\trev(u) &= \trev([v_{i}; v_{j}]) := [v_{j}; v_{i}], \quad \text{and} \\
\trev(t) &= \trev([\zeta_{i}; \tau; \zeta_{j}]) := [\zeta_{j}; \tau; \zeta_{i}] \, .
\end{align}
\vspace{0mm}
An illustration of $\mathbb{T}$ is shown in \cref{fig:loss_viz}. Note that
$\trev$ does not reverse the actual video, \ie,~time does not flow backwards,
but only changes the order in which events happen in the stitched clips. Our
objective is to train a model that is able to distinguish between the original
pair $(u, t)$ and time-reversed versions $(u, \trev(t))$, and $(\trev(u), t)$.

\paragraph{Loss function.}
We assume access to an existing pre-trained video-language model with a visual
encoder $f_{\theta}$ and text encoder $g_{\phi}$. We obtain the video encoding
$\bz_{u} := f_{\theta}(u) \in \mathbb{R}^{d}$ and the text encoding $\bz_{t} :=
g_{\phi}(t) \in \mathbb{R}^{d}$.  Our goal is to adapt $\Theta = \{\theta,
\phi\}$ via post-pretraining such that the resulting model is time aware while
maintaining its original performance on tasks such as retrieval. As we aim to
use a small amount of data, we only update some parameters of the model
($\Theta$), such as the last few layers.

We now introduce our recipe for temporal adaptation based on the InfoNCE
loss~\cite{infonce} to learn time-order sensitive video-text correspondence. For
a positive (or time-order consistent) video-text pair $(u, t)$, we first define
a forward loss where the stitched pair is in its original time-order.
\begin{equation}
\mcL_f = \sum_{(u, t) \in \mathcal{B}}
\left(
\tnce(\bz_{u}, \bz_{t}) + \tnce(\bz_{t}, \bz_{u})
\right),
\end{equation}
where $\tnce$ is the Noise Contrastive Estimation 
(NCE) loss for temporal adaptation, defined as:
\begin{equation}
\label{eq:tnce}
\tnce(\bz_{u}, \bz_{t}) := - \log \frac{
\exp (\bz_{u} \cdot \bz_{t})
}
{
\sum_{t' \in \mathcal{B}_{t} }\exp (\bz_{u} \cdot \bz_{t'})
+ \mcC^{\text{time}}
} \, ,
\end{equation}
where $\mcB$ is the batch of $(u, t)$ pairs and $\mcB_t$ specifically refers to
other stitched text captions in the batch. $\mcC^{\text{time}}$ accumulates
negatives defined using time-order reversal as:
\begin{equation}
\Ctime = 
\alphasame \exp (\bz_{u} \cdot \bz_{\trev(t)}) + 
\alphacross \!\!\!\! \sum_{t' \in \mathcal{B}_{t} \setminus \{t\}}
\!\!\!\! \exp (\bz_{u} \cdot \bz_{\trev(t')}) ,
\end{equation}
where $\alphasame$ controls the effect of contrasting text from the same sample
but with reversed text time-order, \ie,~$\trev(t)$, and $\alphacross$ encourages
the model to contrast between reversed versions of other text captions,
\ie,~$\trev(t')$. Note that when both $\alphasame$ and $\alphacross$ are 0, we
revert back to the standard NCE formulation, albeit on stitched pairs. While
Eq.~\eqref{eq:tnce} corresponds to the video-text loss $\tnce(\bz_u, \bz_t)$,
the text-video loss $\tnce(\bz_t, \bz_u)$ is defined symmetrically. Furthermore,
we also apply a reverse loss $\mcL_{r}$ to bring time-order reversed versions of
both the video and the text together. Note that as we consider $(u, t)$ as a
positive pair, $(\trev(u), \trev(t))$ also form a positive pair,
\begin{equation}
\mcL_r = \!\!\!\!\!\!\!\! \sum_{(\trev(u), \trev(t)) \in \mathcal{B}} \!\!\!\!\!\!\!\!
\left(
\tnce(\bz_{\trev(u)}, \bz_{\trev(t)}) + \tnce(\bz_{\trev(t)}, \bz_{\trev(u)})
\right) \, .
\end{equation}
Here, the $\tnce$ term operates on time-reversed clips and $\mcC^{\text{time}}$
contrasts $(\trev(u), \trev(t))$ with un-reversed text clips in the batch
$(\trev(u), t)$.
The overall loss function is defined as:
\begin{equation}
\mcL = \mcL_f + \beta \mcL_r \, .
\end{equation}
Depending on the choice of loss coefficients $\alphasame, \alphacross, \beta \in
\{0,1\}$, we can vary properties of the adapted model. For example, setting
$\alphasame {=} 1$ encourages high sensitivity to time-order reversal. As we
will see empirically, the loss coefficients also provide the flexibility to
adapt the model to various datasets.

We illustrate this temporal extension of the contrastive loss in
\cref{fig:loss_viz} (best seen in colour). $\mathbb{T}$ illustrates the time
order reversal function. The top half corresponds to $\mcL_f$ while the bottom
half visualizes $\mcL_r$. In particular, the top-left quadrant alone corresponds
to the standard contrastive loss on stitched pairs. While the green diagonal
terms are positive pairs, the red diagonal terms are the strongest drivers for
instilling temporal understanding in the model.

%% file: sections/expts-adaptation.tex
\section{Experiments: TACT Ablations}
\label{sec:ablations}

\paragraph{Base model.}
We demonstrate the effectiveness of TACT as an adaptation recipe on top of
VideoCLIP~\cite{xu-etal-2021-videoclip} owing to its simple architecture,
contrastive objective, and use of pre-computed S3D~\cite{s3d} features that
enable faster experimentation and allow encoding a long temporal context
($\sim$32 secs). We initialize $\Theta$ from the model pretrained on
HowTo100M~\cite{miech19-howto100m} and post-pretrain on multiple datasets.

\paragraph{Datasets.}
One of our key objectives is to post-pretrain on a small amount of data in
contrast to massive pretraining datasets such as WebVid2M~\cite{webvid-Bain21}
or HowTo100M~\cite{miech19-howto100m}. We consider dense video captioning
datasets that offer sufficient diversity in terms of size, backgrounds, clip
durations, viewpoints and activities. Specifically, we experiment with:
(i)~\textit{\textbf{TEMPO}}~\cite{hendricks-etal-2018-localizing-tempo}: the
subset of stitched diverse third-person videos from
DiDeMo~\cite{didemo-Hendricks2017LocalizingMI} with text descriptions for fixed
5s segments that contain before/after relations;
(ii)~\textit{\textbf{ActivityNet Captions}}~\cite{krishna2017dense}: a dense
video captioning dataset with human-centric actions;
(iii)~\textit{\textbf{Charades}}~\cite{charades}: a scripted indoor daily human
activities  video dataset; and
(iv)~\textit{\textbf{Charades-Ego}}~\cite{charades-ego}: similar to Charades,
scripted human activities from the egocentric viewpoint. To construct stitched
clips, we randomly sample any two non-overlapping clip-text pairs in the video.
Since we require stitched clips instead of raw videos, we create new splits for
each dataset (see \cref{tab:dataset-stats}).

\input{tables/2_dataset-stats}

\paragraph{Evaluation metrics.}
We evaluate the post-pretrained model using two sets of metrics: (i)~standard
retrieval metrics, recall $R@1, R@5, R@10$ and median-rank evaluated on stitched
video-text clips; and (ii)~time-order consistency, \ie, the fraction of videos
for which the model correctly associates text that is time order consistent with
the video:
\begin{equation}
\Atime := \frac{1}{|\mcD|} \sum_{(u, t) \in \mcD}
\mathbb{I}[d(\bz_{u}, \bz_{t}) < d(\bz_{u}, \bz_{\trev(t)})],
\label{eq:temporal-acc}
\end{equation}
where $(u, t)$ are time-order consistent pairs, $\mcD$ is the dataset, and
$d(\cdot, \cdot)$ is a distance metric based on cosine similarity.

\paragraph{Post-pretraining details.}
We freeze the word embeddings and layers 1 to 5 for both the video and text
encoders in VideoCLIP. For adaptation, we use the Adam
optimizer~\cite{Adam-kingma} with learning rate $5e^{-6}$, batch size $32$
trained on a single node with 4 GeForce GTX 1080 GPUs. On TEMPO, we train for 60
epochs while on the other datasets, we train for 10 epochs and pick the
checkpoint that maximizes the geometric mean of $R@1$ and $\Atime$ on the
respective validation set. A typical adaptation run takes about 1-3 hours.

\input{tables/4_tact-testset}
\input{tables/3_tact-losscoeff}

\paragraph{Evaluation on the test set.}
Results in \cref{tab:tact-test-results} show that TACT$^\star$ with optimal loss
coefficients outperforms TACT$^\dagger$ (all 0 loss coefficients) and the
zero-shot baseline (no post-pretraining similar to the synthetic data
experiment), both on the retrieval and time-order consistency tasks. This
indicates the robustness of the adaptation.

\paragraph{Impact of loss coefficients.}
Choosing appropriate values for loss coefficients $\Theta_{l} {:=} \{\alphasame,
\alphacross, \beta\}$ allows the model to learn various aspects and adapt using
different datasets. On each dataset, we vary $\Theta_{l} {\in} \{0, 1\}^{3}$ and
find the best configuration based on the $\operatorname{GeometricMean}(R@1,
\max(\Atime - 50, 0))$ on the validation sets. The above metric ensures the
geometric mean is not overpowered by $R@1$. The results are shown in
\cref{tab:temporal-adaptation}. As $\alphasame$ is directly responsible for
discriminating between original and time-reversed orders, unsurprisingly,
setting it to 1 is necessary to achieve the best results on $\Atime$ for all the
datasets. For TEMPO and Charades-Ego, using all loss components (all 1) provides
the best results, whereas $\alphacross {=} 1$ and $\beta {=} 0$ achieves a
better trade-off for ActivityNet and Charades. Choosing $\beta{=}1$ leads to an
improvement in retrieval performance for TEMPO and Charades-Ego but leads to a
decline for ActivityNet and Charades. We attribute this to the number of unique
videos in the train set for these datasets. As ActivityNet and Charades have
more videos than TEMPO or Charades-Ego (see train $N_\mcV$
\cref{tab:dataset-stats}) additional positives introduced by setting $\beta{=}1$
are not necessary and in fact, hurt performance. Finally, we note that carefully
setting the value of $\alphasame$ provides a convenient trade-off between
spatial and temporal understanding. Please refer to \cref{para:spatial-vs-temporal} for detailed
experiments.

\begin{figure}[t!]
\captionsetup{skip=2mm,font=small}
\centering
\includegraphics[width=\linewidth]{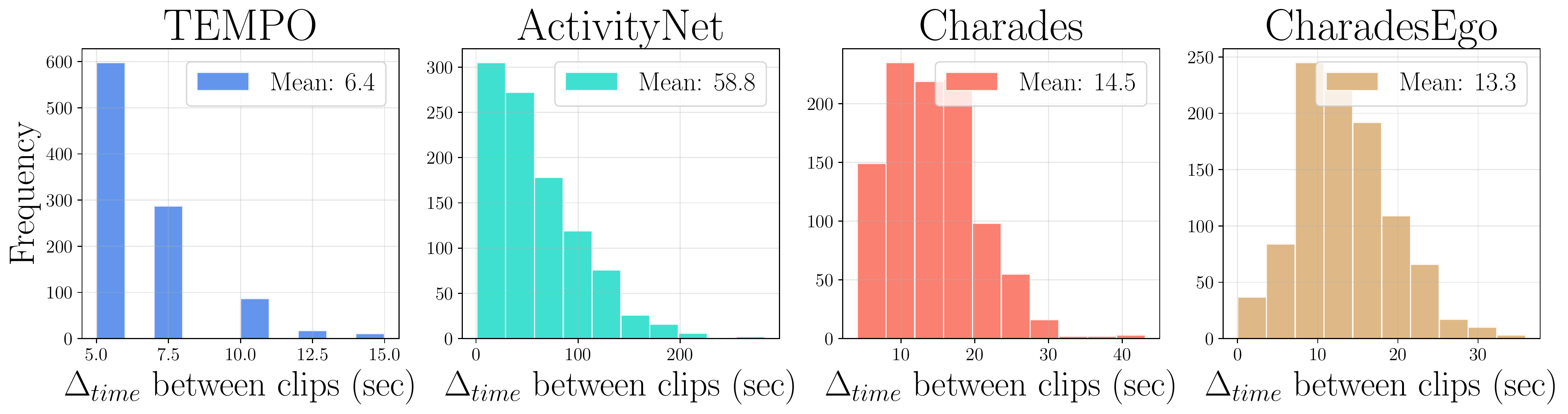}
\caption{Time-distance between stitched clips in datasets for temporal
adaptation ($\Deltatime$). TEMPO has stitched clips close to each other while
those in Charades-Ego are farthest apart suggesting a correlation between
$\Delta_{\text{time}}$ and  difficulty of temporal adaptation.}
\label{fig:time-distance}
\end{figure}

\paragraph{What makes temporal adaptation hard?} We observe a large gap in
$\Atime$ between TEMPO and ActivityNet. We hypothesize that the distance (in
seconds) between the two clips ($\Deltatime$) in a stitched clip is strongly
correlated with the difficulty of adaptation. Intuitively, it is easier to infer
the time-order consistency for a stitched clip $u$ with text $t$ that has
distant constituent clips $v_i, v_j$ since the objects and scene could be vastly
different. In contrast, it is harder to discern the correct time-order when the
constituent clips are closer in time. \cref{fig:time-distance} shows
distribution of $\Deltatime$ for each dataset. Indeed, $\Deltatime$ in
ActivityNet (58.8s) is much higher than that in TEMPO (6.4s) making the task
harder on TEMPO. To further test our hypothesis, we conduct a controlled
experiment where we gradually vary the distribution of $\Deltatime$ for
Charades-Ego to match it to that of TEMPO. We find a strong correlation ($\rho
{=} 0.92$) between $\Deltatime$ and hardness of adaptation. Please refer to
\cref{para:temporal-adaptation-hardness} for more details.

%% file: tables/2_dataset-stats.tex
\begin{table}[t]
\captionsetup{skip=2mm,font=small}
\centering
\tabcolsep=0.09cm
\resizebox{1\linewidth}{!}{
\begin{tabular}{l rrrrrr cr}
\toprule
\multicolumn{1}{l}{\textbf{Dataset}} & \multicolumn{2}{c}{Train} & \multicolumn{2}{c}{Validation} & \multicolumn{2}{c}{Test} & Ego & Length\\ \cmidrule{2-7}
 & \multicolumn{1}{c}{$N_{\mcV}$} & \multicolumn{1}{c}{$N_{\mcC}$} & \multicolumn{1}{c}{$N_{\mcV}$} & \multicolumn{1}{c}{$N_{\mcC}$} & \multicolumn{1}{c}{$N_{\mcV}$} & \multicolumn{1}{c}{$N_{\mcC}$} & & (s) \\
 \midrule
\vspace{1mm}
\cellcolor[HTML]{E4E6FF}TEMPO & 3,904 & 28,427 & 411 & 1,000 & 396 & 1,000  & \xmark &  30 \\
\vspace{1mm}
\cellcolor[HTML]{DEFEFD}ActivityNet & 7,440 & 95,474 & 453 & 906 & 456 & 912 & \xmark & 120 \\
\vspace{1mm}
\cellcolor[HTML]{FEE1DF}Charades & 5,262 & 99,928 & 500 & 1,000 & 500 & 1,000 & \xmark & 30 \\
\vspace{1mm}
\cellcolor[HTML]{F8E8C6}Charades-Ego & 2,679 & 155,306 & 500 & 1,000 & 210 & 420 & \cmark & 31 \\
\bottomrule
\end{tabular}%
}
\caption{Statistics of datasets we consider for temporal adaptation. $N_{\mcV}$ is the number of unique videos and $N_{\mcC}$ is the number of stitched clips.
Based on $N_{\mcV}$, TEMPO and Charades-Ego are smaller as compared to ActivityNet and Charades.}
\label{tab:dataset-stats}
\end{table}

%% file: tables/4_tact-testset.tex
\begin{table}[b]
\centering
\small
\tabcolsep=0.13cm
\captionsetup{skip=2mm,font=small}
\begin{tabular}{ll rrr}
\toprule
\textbf{Dataset} & \textbf{Method} & \multicolumn{2}{c}{\textbf{Retrieval}} & \multicolumn{1}{c}{\textbf{Time-order}} \\
\cmidrule{2-5} 
 &  & \multicolumn{1}{r}{R@1$ {\uparrow}$} & \multicolumn{1}{r}{MedR $\downarrow$} & \multicolumn{1}{r}{$\Atime {\uparrow}$} \\
 \lgraymidrule
 \cellcolor[HTML]{E4E6FF} & Zero-shot & 3.7 & 49.0 & 48.1 \\
  \cellcolor[HTML]{E4E6FF}TEMPO & TACT$^\dagger$ & 7.7 & 13.0 & 46.5 \\
  \cellcolor[HTML]{E4E6FF} & TACT$^\star$ & \textbf{9.3} & \textbf{9.0} & \textbf{66.5} \\ \lgraymidrule
\multirow{3}{*}{\cellcolor[HTML]{DEFEFD}} & Zero-shot & 1.1 & 44.0 & 49.6 \\
\cellcolor[HTML]{DEFEFD}ActivityNet & TACT$^\dagger$ & \textbf{5.8} & \textbf{34.0} & 59.7 \\
\cellcolor[HTML]{DEFEFD} & TACT$^\star$ & \textbf{5.8} & 35.0 & \textbf{85.7} \\
\lgraymidrule
\multirow{3}{*}{\cellcolor[HTML]{FEE1DF}} & Zero-shot & 1.3 & 170.0 & 47.1 \\
\cellcolor[HTML]{FEE1DF}Charades &  TACT$^\dagger$ & 5.3 & 38.5 & 73.5 \\
\cellcolor[HTML]{FEE1DF} & TACT$^\star$ & \textbf{5.7} & \textbf{35.0} & \textbf{77.0} \\
\lgraymidrule
\multirow{3}{*}{\cellcolor[HTML]{F8E8C6}} & Zero-shot & 1.6 & 64.0 & 53.7 \\
\cellcolor[HTML]{F8E8C6}Charades-Ego & TACT$^\dagger$  & 6.4 & 35.0 & 60.1 \\
\cellcolor[HTML]{F8E8C6} & TACT$^\star$ & \textbf{10.1} & \textbf{28.5} & \textbf{68.2} \\
\bottomrule
\end{tabular}%
\caption{Results for the best TACT model on test sets.
TACT$^\star$ has optimal loss coefficients and TACT$^\dagger$ is a baseline with all coefficients 0.
On time order, TACT generalizes well with TACT$^\star$ outperforming the baselines.
On retrieval, for TEMPO and Charades-Ego, TACT$^\star$ outperforms
the baseline as their optimal models have $\beta{=}1$ which helps retrieval with a small amount of data.
}
\label{tab:tact-test-results}
\end{table}

%% file: tables/3_tact-losscoeff.tex
\begin{table*}[t!]
\captionsetup{skip=2mm,font=small}
\small
\centering
\tabcolsep=0.12cm
\begin{tabular}{ccc rrr c rrr c rrr c rrr}
\toprule
\multicolumn{3}{c}{\textbf{Loss coefficients}} & \multicolumn{3}{c}{\cellcolor[HTML]{E4E6FF}\textbf{TEMPO}} & & \multicolumn{3}{c}{\cellcolor[HTML]{DEFEFD}\textbf{ActivityNet}} & & \multicolumn{3}{c}{\cellcolor[HTML]{FEE1DF}\textbf{Charades}} & & \multicolumn{3}{c}{\cellcolor[HTML]{F8E8C6}\textbf{Charades-Ego}} \\ \midrule
$\alpha_{\text{same}}$ & $\alpha_{\text{cross}}$ & $\beta$ & R@1 $\uparrow$ & MedR $\downarrow$ & $A_{\text{time}} \uparrow$ & & R@1 $\uparrow$ & MedR $\downarrow$ & $A_{\text{time}} \uparrow$ & & R@1 $\uparrow$ & MedR $\downarrow$ & $A_{\text{time}} \uparrow$ & & R@1 $\uparrow$ & MedR $\downarrow$ & $A_{\text{time}} \uparrow$ \\
\midrule
 & {\small Chance} &  & 0.1 & 500.0 & 50.0 & & 0.1 & 453.0 & 50.0 & & 0.1 & 500.0 & 50.0 & & 0.1 & 500.0 & 50.0 \\
\graymidrule
0 & 0 & 0 &
8.3	& 14.0 & 49.4 & &
6.4	& 30.0 & 57.3 & &
5.7 & 42.0 & 71.5 & &
2.9 & 44.0 & 64.6 \\
0 & 0 & 1 &
8.2 & 14.0 & 49.5 & &
5.6	& 27.0 & 47.0 & &
4.2 & 58.0 & 75.1 & &
3.2 & 41.5 & 65.2 \\
0 & 1 & 0 &
8.2	& 15.0 & 49.3 & &
6.1	& 29.0 & 78.8 & &
5.2 & 45.0 & 78.9 & &
3.4 & 38.0 & 64.5 \\
0 & 1 & 1 &
8.1	& 14.0 & 49.5 & &
5.8 & 27.0 & 48.3 & &
4.2 & 58.0 & 75.1 & &
3.1 & 41.0 & 67.0 \\
1 & 0 & 0 &
6.4	& 20.0 & 60.6 & &
5.9 & 28.0 & 79.1 & &
6.1 & 38.0 & 76.3 & &
3.2 & 42.0 & 66.1 \\
1 & 0 & 1 &
6.5	& 24.0 & 62.9 & &
5.6 & 26.0 & 63.1 & &
4.9 & 51.0 & 78.0 & &
3.3 & 39.0 & 70.7 \\
1 & 1 & 0 &
5.9 & 24.0 & 59.7 & &
\cellcolor[HTML]{DEFEFD}6.0 & \cellcolor[HTML]{DEFEFD}29.0 & \cellcolor[HTML]{DEFEFD}86.3 & &
\cellcolor[HTML]{FEE1DF}6.6 & \cellcolor[HTML]{FEE1DF}43.0 & \cellcolor[HTML]{FEE1DF}77.8 & &
3.7 & 40.5 & 67.9 \\
1 & 1 & 1 &
\cellcolor[HTML]{E4E6FF}7.5 & \cellcolor[HTML]{E4E6FF}17.0 & \cellcolor[HTML]{E4E6FF}62.5 & &
5.7 & 27.0 & 63.8 & &
5.1 & 51.0 & 77.7 & &
\cellcolor[HTML]{F8E8C6}3.8 & \cellcolor[HTML]{F8E8C6}38.5 & \cellcolor[HTML]{F8E8C6}68.3 \\
\bottomrule
\end{tabular}%
\caption{Impact of loss coefficients for TACT post-pretraining on validation sets of various datasets.
For each dataset, the corresponding \emph{color-marked row denotes the best configuration} based on the geometric mean of $R@1$ and $\Atime$. TACT is able to connect time-order in video and language while maintaining its retrieval capabilities across several datasets.
}
\label{tab:temporal-adaptation}
\end{table*}

%% file: sections/expts-downstream.tex
\section{Experiments: Downstream Evaluation}
\label{sec:downstream}

\input{tables/5_downstream-zero-shot}

The goal of video-language foundation models is to generalize in a zero- or
few-shot manner to a diverse range of downstream tasks. We evaluate TACT models
on three sets of downstream tasks that need low-to-high  time awareness.

\paragraph{Baseline: Standard post-pretraining.}
Comparing our temporally adapted models with pretrained VideoCLIP is not fair
since adapted models see data beyond the pretraining phase. In addition to the
zero-shot comparison, we compare against a baseline model that is trained for
standard video-text retrieval on the same datasets as temporal adaptation.
Instead of using stitched clips, we use simple canonical pairs, \ie,~$(v_i,
\zeta_i)$ instead of $(u_{ij}, t_{ij})$.

\paragraph{Evaluating TACT adapted models on synthetic data.}
On the video-to-text variant, TACT adapted on TEMPO achieves $64.4\%$,
ActivityNet $52.5\%$, Charades $65.0\%$, Charades-Ego $85.6\%$. This is usually
higher than the performance that non-adapted models achieve in
\cref{tab:synthetic-v2t}. This highlights that TACT models learn useful signals
to match  time-order in video and language.

\paragraph{I. Text-to-video retrieval.} 
We consider two widely used benchmarks:
\textit{\textbf{MSR-VTT}}~\cite{xu2016-msr-vtt} and
\textit{\textbf{YouCookII}}~\cite{ZhXuCoAAAI18-youcook2} and adopt standard
retrieval metrics. Recent work has identified a bias for spatial understanding
in these datasets~\cite{buch2022-revisiting, revealing-single-frame,
8578867-what-makes-video,clip-H,Luo2021-CLIP4Clip, Li2022BLIPBL}. Thus, we
consider this class of tasks as requiring low time awareness. As shown in
\cref{tab:downstream-full} set~I, on MSR-VTT~\cite{xu2016-msr-vtt}, we observe
that TACT is worse (marked in \marktext{red!15}{red}) or at par with the
baseline across datasets. This aligns well with findings
in~\cite{revealing-single-frame,buch2022-revisiting} that these benchmarks do
not need time awareness. On YouCookII~\cite{ZhXuCoAAAI18-youcook2}, TACT models
based on Charades(-Ego) outperform the baseline (marked in
\marktext{green!15}{green}). We believe this is a consequence of a lower domain
shift between YouCookII and Charades.

\paragraph{II. Temporal video QA.}
Next, we use subsets of recently released multiple-choice video question
answering benchmarks:  \textit{\textbf{Next-QA}}~\cite{xiao2021-nextqa} and
\textit{\textbf{AGQA}}~\cite{GrundeMcLaughlin2021-AGQA}. The idea is to check if
we can probe models for temporal understanding by asking questions with temporal
language. Buch~\etal~\cite{buch2022-revisiting} identify a subset of Next-QA,
dubbed as \texttt{ATP-hard}\footnote{Available here:
\url{github.com/StanfordVL/atp-video-language}}, with a higher concentration of
temporally challenging data. For AGQA, we pick a subset of $\sim$6k questions
that explicitly have a question with before/after relation. We consider these
benchmarks as requiring a moderate-high level of time awareness and AGQA in
particular is also close to our adaptation task. We use accuracy as the standard
metric. We observe (see \cref{tab:downstream-full} set~II) that indeed TACT
almost always outperforms (marked in \marktext{green!15}{green}) baselines on
both Next-QA and AGQA. TEMPO-adapted TACT seems to generalize particularly well
on both benchmarks. Likewise, Charades-adapted TACT does well on AGQA since AGQA
is also based on the Charades videos accounting for reduced domain shift. We
affirm that temporal adaptation is useful, especially when the downstream tasks
require it.

\paragraph{III. Action-to-video retrieval.}
\label{para:task-III}
Finally, we consider action recognition benchmarks such as
\textit{\textbf{Something-Something}} (SSv2)~\cite{SS-v2-arxiv} and
\textit{\textbf{Temporal}}~\cite{DBLP:journals/corr/abs-1907-08340-temporal-dataset}.
SSv2 was designed to capture richer temporal
information~\cite{SS-v2-arxiv,revealing-single-frame}. We follow
Lei~\etal~\cite{revealing-single-frame}, who propose the
\textit{template-retrieval} task that encourages temporal modelling and use
their evaluation split\footnote{Available here:
\url{github.com/jayleicn/singularity}} containing $C{=}174$ actions and $K{=}12$
videos per class. Interestingly, different actions in SSv2 require differing
levels of time awareness. We create a subset \textit{\textbf{SSv2 (events)}}
with $C_{\text{events}}{=}49$ actions that have at least two verbs in the label
as the occurrence of multiple verbs is indicative of multiple events occurring
in sequence. Finally, we also evaluate against the Temporal
benchmark~\cite{DBLP:journals/corr/abs-1907-08340-temporal-dataset}, a
combination of 50 action classes from SSv2~\cite{SS-v2-arxiv} and
Kinetics-400~\cite{Kinetics-400-arxiv} for which temporal information is deemed
to be essential for recognition. Similar to text-to-video retrieval, we use the
action class as a text query and obtain a ranking over all videos. Different
from the retrieval setup, since a single query has multiple correct answers (up
to $K{=}12$ videos), we report mAP as the metric for these benchmarks. This task
set needs high time awareness. Furthermore, unlike QA tasks in II, there is a
\textit{shift} in several (uncontrolled) factors as we move from temporal
adaptation task to these tasks. From \cref{tab:downstream-full}, we observe that
TEMPO- and Charades-adapted models generalize well across set III benchmarks.
ActivityNet-adapted TACT underperforms on SSv2 but outperforms the baseline on
strongly temporal actions in SSv2 (events) and Temporal. Finally, TACT adapted
on Charades-Ego is at-par or slightly worse than the baseline on SSv2 variants,
and also on Temporal, perhaps due to the shift from egocentric to third-person
videos. Overall, despite SSv2 and Temporal requiring high time awareness, TACT
models show promising zero-shot generalization with the right choice of the
adaptation dataset.

%% file: tables/5_downstream-zero-shot.tex
\begin{table*}[t!]
\captionsetup{skip=2mm,font=small}
\centering
$\hfill  \text{\small \colorbox{red!15}{Low}} \xrightarrow{{\hspace*{3.6cm}} \text{Time awareness} {\hspace*{3.6cm}}} \text{\small \colorbox{green!15}{High}} {\hspace*{1.2cm}}$
\resizebox{\textwidth}{!}{%
\begin{tabular}{ll rrrrrr c rr c rrr}
\midrule
\multicolumn{2}{c}{\textbf{Adaptation}} & \multicolumn{6}{c}{\textbf{I. Text-to-Video Retrieval}} & & \multicolumn{2}{c}{\textbf{II. Temporal VQA}} & & \multicolumn{3}{c}{\textbf{III. Action-to-Video Retrieval}} \\ \midrule
\textbf{Dataset} & \textbf{Method} & \multicolumn{3}{c}{\textbf{MSR-VTT}} & \multicolumn{3}{c}{\textbf{YouCookII}} & & \multicolumn{1}{c}{\textbf{Next-QA (ATP)}} & \multicolumn{1}{c}{\textbf{AGQA}} & & \multicolumn{1}{c}{\textbf{SSv2}} & \multicolumn{1}{l}{\textbf{SSv2 (events)}} & \multicolumn{1}{c}{\textbf{Temporal}} \\ \midrule
 &  & $R@1$ & $R@5$ & $R@10$ & $R@1$ & $R@5$ & $R@10$ & & Accuracy & Accuracy & & mAP & mAP & mAP \\
 \midrule
- & Chance & 0.1 & 0.5 & 1.0 & 0.1 & 0.5 & 1.0 & & 20.0 & 50.0 & & 0.6 & 2.0 & 2.0 \\
- & None & 10.6 & 23.4 & 29.9 & 18.2 & 45.5 & 59.9 & & 23.4 & 49.9 & & 3.4 & 6.4 & 13.0 \\
\graymidrule
\multirow{2}{*}{TEMPO} & Baseline & 12.0 & 29.3 & 37.3 & 21.5 & 48.2 & 61.8 & & 25.0 & 50.8 & & 3.9 & 6.8 & 15.9 \\
\lgraycmidrule{2-15} 
 & TACT & \cellcolor{green!15}12.8 & \cellcolor{red!15}26.5 & \cellcolor{red!15}35.7 & \cellcolor{red!15}20.4 & \cellcolor{red!15}45.1 & \cellcolor{red!15}58.7 & & \cellcolor{green!15}27.6 & \cellcolor{green!15}57.1 & & \cellcolor{green!15}4.2 & \cellcolor{green!15}7.7 & \cellcolor{green!15}16.1 \\
 \graymidrule
\multirow{2}{*}{ActivityNet} & Baseline & 15.7 & 34.4 & 44.9 & 15.6 & 38.8 & 51.4 & & 23.7 & 50.7 & & 3.7 & 7.0 & 16.0 \\ \lgraycmidrule{2-15} 
 & TACT & \cellcolor{red!15}13.8 & \cellcolor{red!15}29.6 & \cellcolor{red!15}39.6 & \cellcolor{green!15}16.0 & \cellcolor{red!15}36.9 & \cellcolor{red!15}49.8 & & \cellcolor{green!15}25.4 & \cellcolor{red!15}50.3 & & \cellcolor{red!15}3.5 & \cellcolor{green!15}7.2 & \cellcolor{green!15}16.2 \\
 \graymidrule
\multirow{2}{*}{Charades} & Baseline & 12.3 & 25.8 & 33.6 & 21.5 & 48.6 & 61.7 & & 26.0 & 50.5 & & 4.1 & 7.1 & 13.7 \\ \lgraycmidrule{2-15} 
 & TACT & \cellcolor{red!15}11.7 & \cellcolor{red!15}25.2 & \cellcolor{red!15}32.4 & \cellcolor{green!15}22.4 & \cellcolor{green!15}49.1 & \cellcolor{green!15}62.4 & & \cellcolor{red!15}25.2 & \cellcolor{green!15}54.8 & & \cellcolor{green!15}4.3 & \cellcolor{green!15}7.8 & \cellcolor{green!15}14.1 \\
 \graymidrule
\multirow{2}{*}{Charades-Ego} & Baseline & 13.1 & 27.5 & 34.5 & 19.4 & 47.1 & 60.8 & & 24.3 & 49.7 & & 4.0 & 6.9 & 14.7 \\ \lgraycmidrule{2-15} 
 & TACT & \cellcolor{red!15}11.1 & \cellcolor{red!15}24.6 & \cellcolor{red!15}30.6 & \cellcolor{green!15}21.9 & \cellcolor{green!15}48.2 & \cellcolor{green!15}61.9 & & \cellcolor{green!15}25.6 & \cellcolor{green!15}58.4 & & \cellcolor{red!15}3.9 & 6.9 & \cellcolor{red!15}13.5 \\
 \bottomrule
\end{tabular}%
}
\caption{
Results on downstream zero-shot evaluation with tasks requiring increasing time awareness from I to III.
None corresponds to direct evaluation of the VideoCLIP model on the downstream dataset.
\colorbox{green!15}{Green} denotes an improvement for the TACT adapted model w.r.t. the baseline,
\colorbox{red!15}{red} denotes a deterioration.
As we move from tasks that need low to high time awareness, the effectiveness of TACT increases.
See
\cref{para:task-III} for a more detailed discussion.
The table is best viewed on screen in colour.
}
\label{tab:downstream-full}
\end{table*}

%% file: sections/discussion.tex
\section{Discussion and Conclusion}

\begin{figure}[t]
\captionsetup{skip=0mm}
    \centering
    \includegraphics[width=0.9\linewidth]{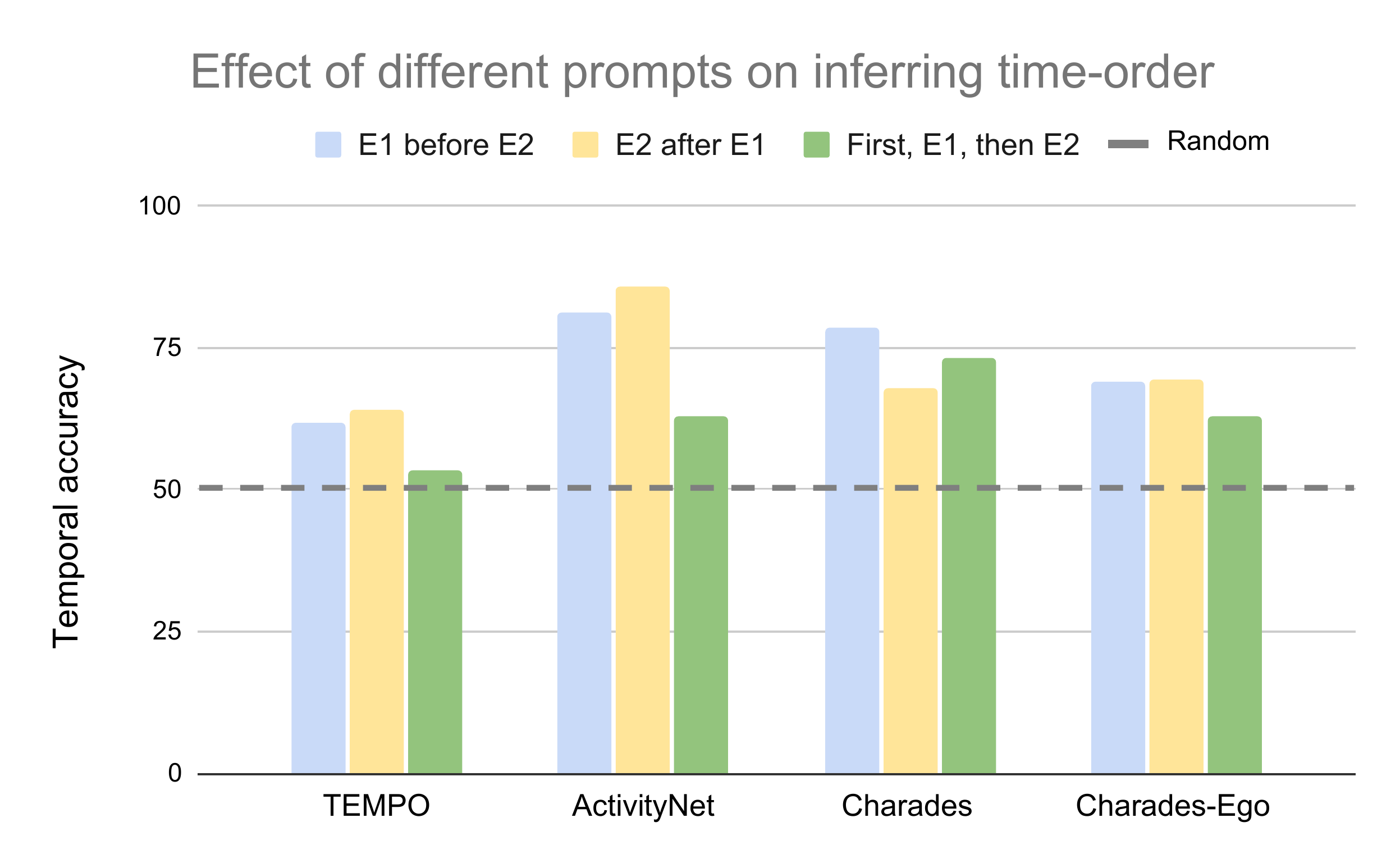}
    \caption{Models trained by TACT with before/after relations generalize to a
        new kind of prompt such as \texttt{First,~\kern-.05em..,
        then~\kern-.05em..} indicating learning of the underlying true
        time-order of events.}
    \label{fig:different-prompts}
\end{figure}

\paragraph{Generalization to other temporal prompts.}
The time-order of events in language can be described using various sentence
structures. Although we train video-language models using before/after
relations, it is natural to ask if the model still correctly associates
time-order for a different prompt such as ~\texttt{First,\kern-.05em..,
then,\kern-.05em..}. To systematically test this, we gather event pairs $E_{1},
E_{2}$ ($E_{1}$ occurs before $E_{2}$ in the video) for each sample in the
validation set and stitch them using three prompts as follows: (i) $E_1$
\texttt{before} $E_2$, (ii) $E_2$ \texttt{after} $E_1$, (iii) \texttt{First}
$E_1$, \texttt{then} $E_2$. As shown in \cref{fig:different-prompts},
TACT-adapted models generalize well to a new prompt (iii). This substantiates
the learning of time-order of events rather than merely learning the order of
words in the sentence.

\paragraph{Limitations.}
While we present a promising way of instilling time in video-language models,
our work is limited to the  VideoCLIP~\cite{xu-etal-2021-videoclip} pretrained
model. Our initial experiments with Frozen in Time~\cite{Bain21-frozen} were not
as promising, perhaps because it uses a much shorter temporal context (4
frames). Please see \cref{para:more-pretrained-models} for results on more pretrained models.
Furthermore, we consider a specific definition of time awareness derived from
temporal relations like before/after. It is natural to ask if this can be
extended to more general notions of temporality, \eg,~as defined by
Allen~\cite{time-in-lang2-Allen1984TowardsAG}. Finally, there can always be more
downstream tasks considered such as (spatio-)temporal localization.

\paragraph{Conclusion.}
Given the essence of time in video-language models, we present a simple
experiment based on synthetic data to test for time awareness. We find that
existing models lack a sense of time defined in terms of consistency of order of
events in video and language. To fill this gap, building upon
VideoCLIP~\cite{xu-etal-2021-videoclip}, we present TACT, a recipe to instill
this sense of time in video-language models. Finally, we analyze the zero-shot
generalizability of TACT-adapted models to a diverse set of tasks. We hope that
this work provokes further probing and instilling time awareness in
video-language models, and also inspires other adaptations of foundational
models to solve various challenging tasks.

\paragraph{Acknowledgements.} We acknowledge support from ELLIS Amsterdam and
the AMS Scholarship to Piyush. We thank Dr Dennis Koelma for their help with
compute infrastructure and Dr Hazel Doughty for useful discussions.

%% file: sections/supp.tex
\appendix

\section*{Supplementary Material}

As part of the supplementary material, we describe pre-processing steps as well
as some qualitative examples from the datasets in \cref{sec:datasets}. In
\cref{sec:supp-expts}, we present additional ablations on what makes temporal
adaptation hard. This expands on the last paragraph of \cref{sec:ablations} of the main paper.
Finally, in \cref{sec:qualitative-analysis}, we conduct a qualitative analysis
to verify if the model has indeed learnt to connect the time order.

\section{Datasets and Pre-processing}
\label{sec:datasets}

We sketch out the procedure we use for stitching two clips within a video.

\begin{figure}[t]
\captionsetup{skip=2mm,font=small}
\centering
\includegraphics[width=\columnwidth]{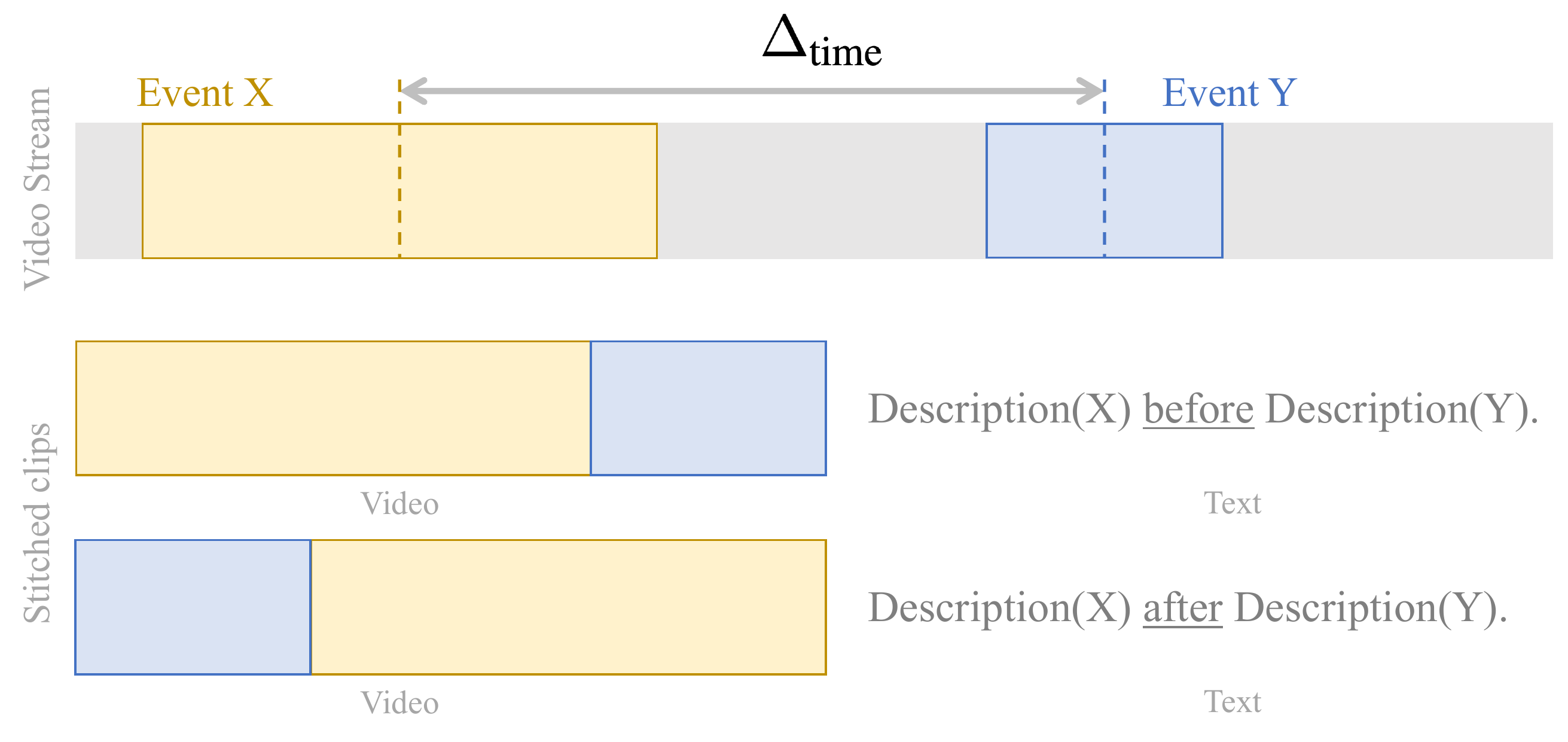}
\caption{Illustration of clip stitching. We consider two non-overlapping events
in a video and stitch them with temporal relations - \texttt{before} and
\texttt{after}. $\Deltatime$ denotes the time difference between midpoints of
the two events.
}
\label{fig:stitching}
\end{figure}

\paragraph{Clip stitching.} Consider a video containing two events (clips)
\colorbox{yellow!30}{$v_{i}$}, \colorbox{myblue!15}{$v_{j}$} with associated
captions $\zeta_{i}, \zeta_{j}$ as shown in \cref{fig:stitching}. We assume
these are non-overlapping (in time). We stitch the text descriptions to
construct a new caption $t_{ij} := [\zeta_{i}; \tau; \zeta_{j}]$. Since $\tau$
can be either \texttt{before} or \texttt{after}, we end up with two newly
constructed sentences. Corresponding to each of these new sentences, we also
stitch the video events to construct a stitched video. Note that the order of
stitching video events depends on the value of $\tau$. For example, if $\tau$ is
\texttt{before}, then $u_{ij} := [v_{i};v_{j}]$ as shown in first of the two
stitched clips. If $\tau$ is \texttt{after}, then $u_{ij} := [v_{j}; v_{i}]$ as
shown in the second of the two stitched clips.

From each stitched clip in \cref{fig:stitching}, we construct negatives for the
contrastive loss by reversing the time order in either video or text. This step
happens on-the-fly during loss computation, and hence, we do not show it here.
For a given dataset, we can either use all possible tuples of non-overlapping
events to create such stitched clips or sample from all possible tuples. Since
the TEMPO dataset already comes with stitched event descriptions (based on
DiDeMo), we directly use its subset which has \texttt{before}/\texttt{after}
relations in the text. For all the other datasets, we apply the stitching
process as described. Recall, $\Deltatime$ is the time distance between the two
events, and plays a key role in deciding the difficulty of temporal adaptation,
as observed empirically.

Next, we describe dataset properties and show some qualitative examples after
the clip stitching step.

\begin{figure}[ht]
\captionsetup{skip=2mm,font=small}
\centering
\begin{subfigure}[b]{\columnwidth}
   \includegraphics[width=1\linewidth]{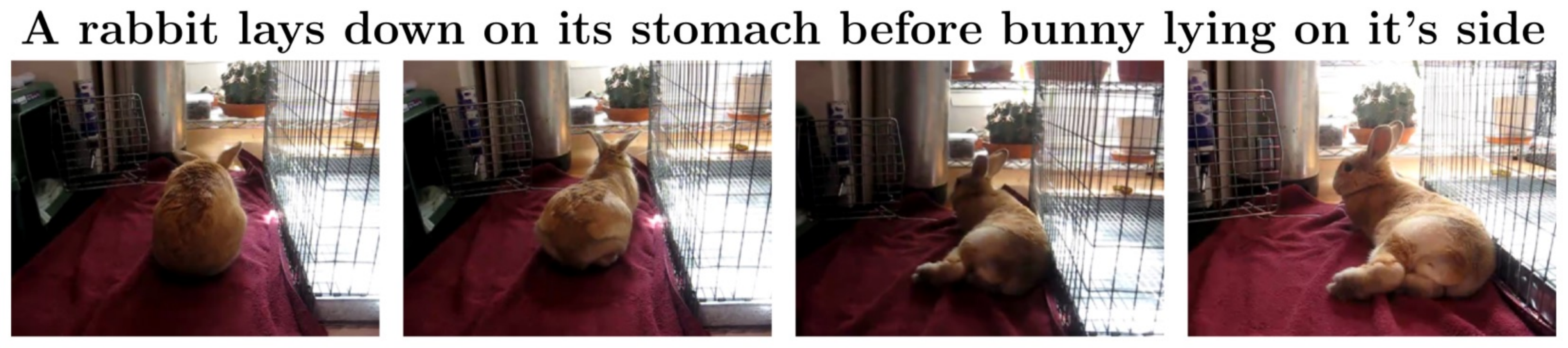}
\end{subfigure}

\begin{subfigure}[b]{\columnwidth}
   \includegraphics[width=1\linewidth]{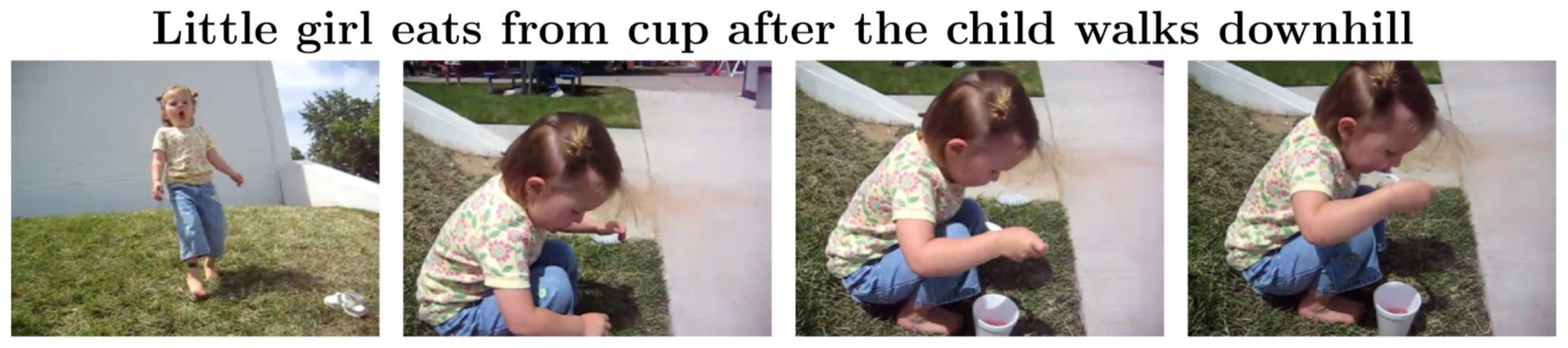}
   \caption{TEMPO}
   \label{fig:tempo-2}
\end{subfigure}

\begin{subfigure}[b]{\columnwidth}
   \includegraphics[width=1\linewidth]{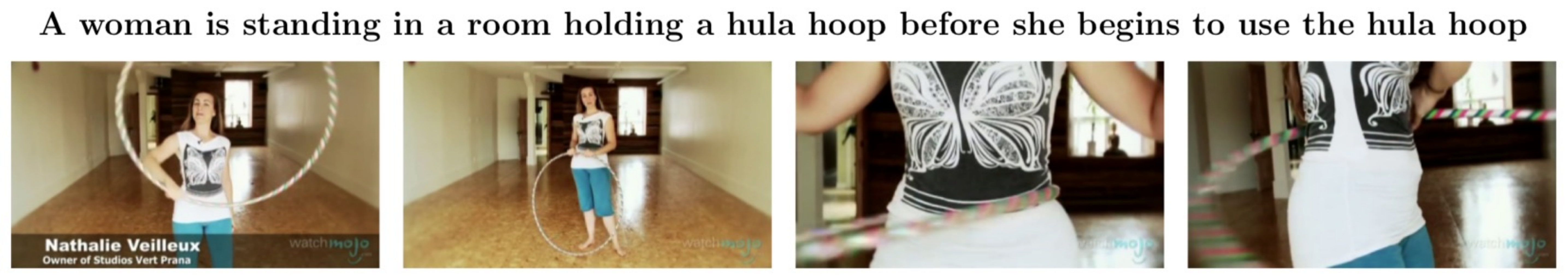}
\end{subfigure}

\begin{subfigure}[b]{\columnwidth}
   \includegraphics[width=1\linewidth]{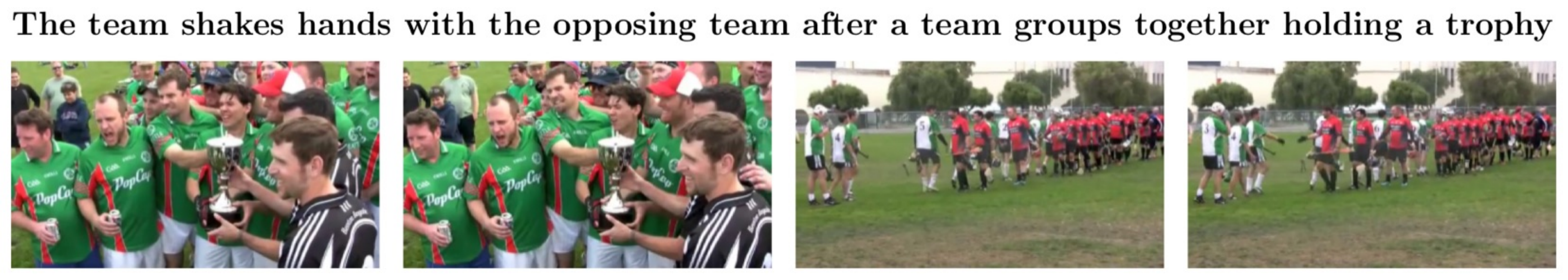}
   \caption{ActivityNet}
   \label{fig:anet-2}
\end{subfigure}

\begin{subfigure}[b]{\columnwidth}
   \includegraphics[width=1\linewidth]{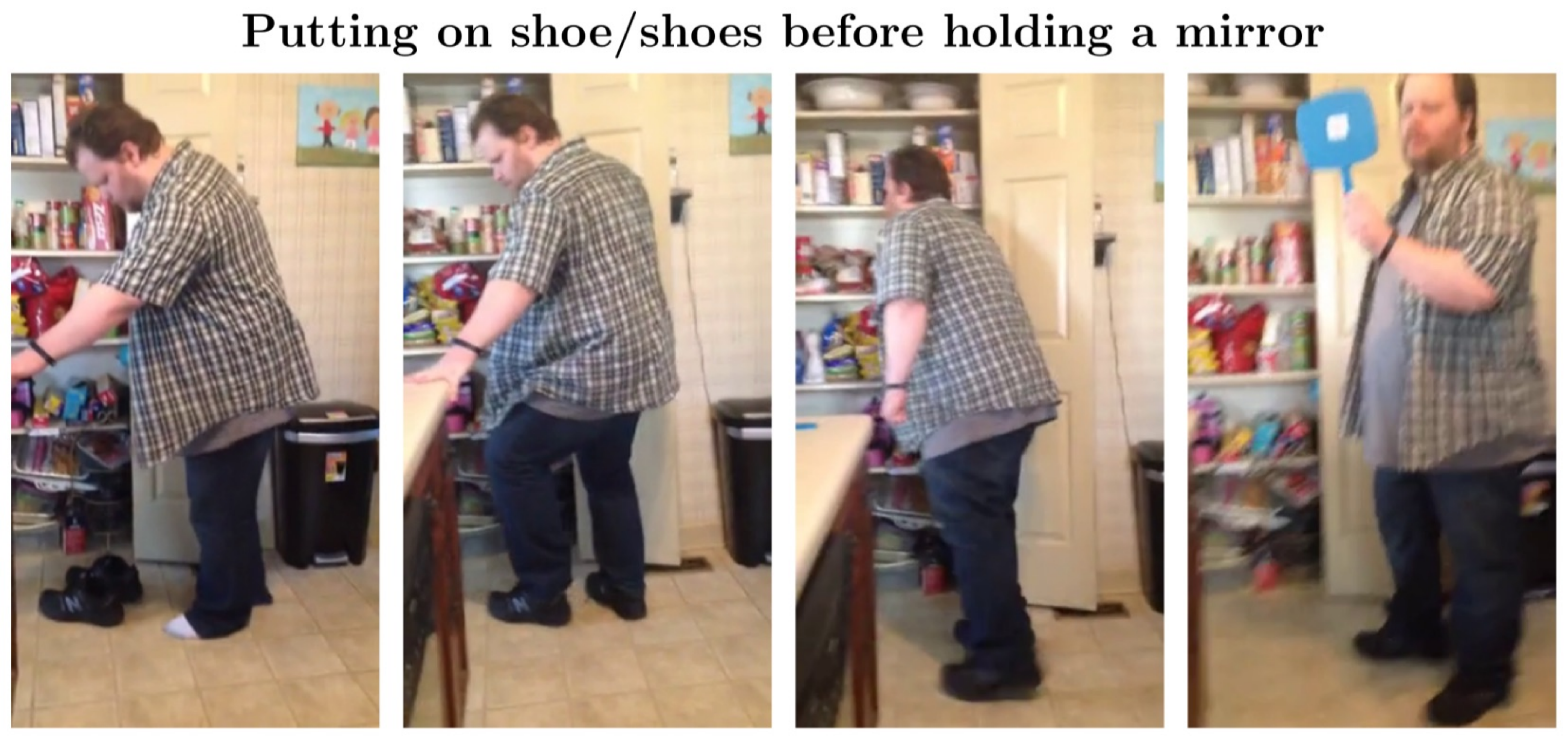}
   \caption{Charades}
   \label{fig:charades-2}
\end{subfigure}

\begin{subfigure}[b]{\columnwidth}
   \includegraphics[width=1\linewidth]{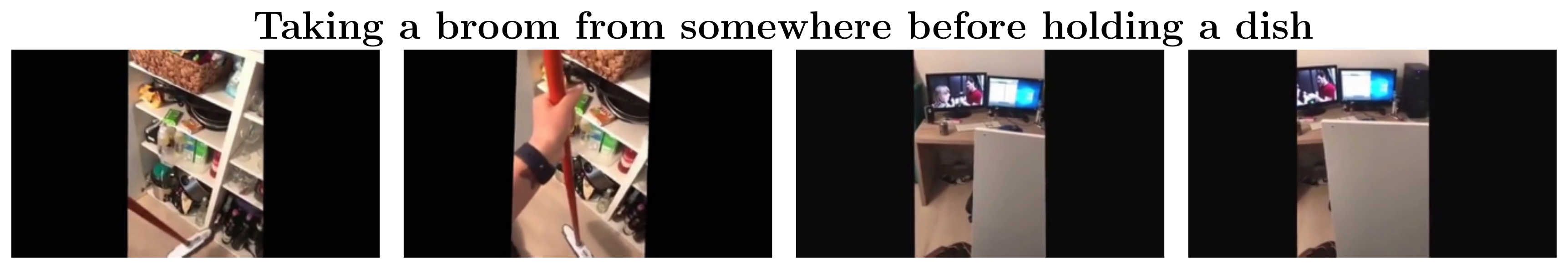}
   \caption{Charades-Ego}
   \label{fig:charadesego-2}
\end{subfigure}

\caption{Examples from datasets used for temporal adaptation. The first two
   frames are linearly spaced from the first event while the last two from the
   second event. Notice how there is a significant change in visual context
   between the two events in ActivityNet in contrast to other datasets. Best
   viewed on a screen.}
\label{fig:examples}
\end{figure}

\paragraph{Adaptation datasets.}
To gain a sense of the diversity in the datasets we consider for adaptation, we
present examples of stitched clips from these datasets in \cref{fig:examples}.
Since TEMPO has short adjacent clips, the context remains almost the same, we
think this is important to instill a sense of time in models. In contrast, for
ActivityNet, since the stitched events are far apart, the context changes make
it easy to infer which event description goes with which part of the video, or
the time order of events. In this regard, Charades and Charades-Ego are similar
to TEMPO. Quantitatively, this change in context is captured by $\Deltatime$
which is lowest for TEMPO (mean 6.8s), followed by Charades-Ego (13.3s),
Charades (14.5s) and ActivityNet (58.8s). 

\begin{figure}[t]
\captionsetup{skip=2mm}
    \centering
    \includegraphics[width=\columnwidth]{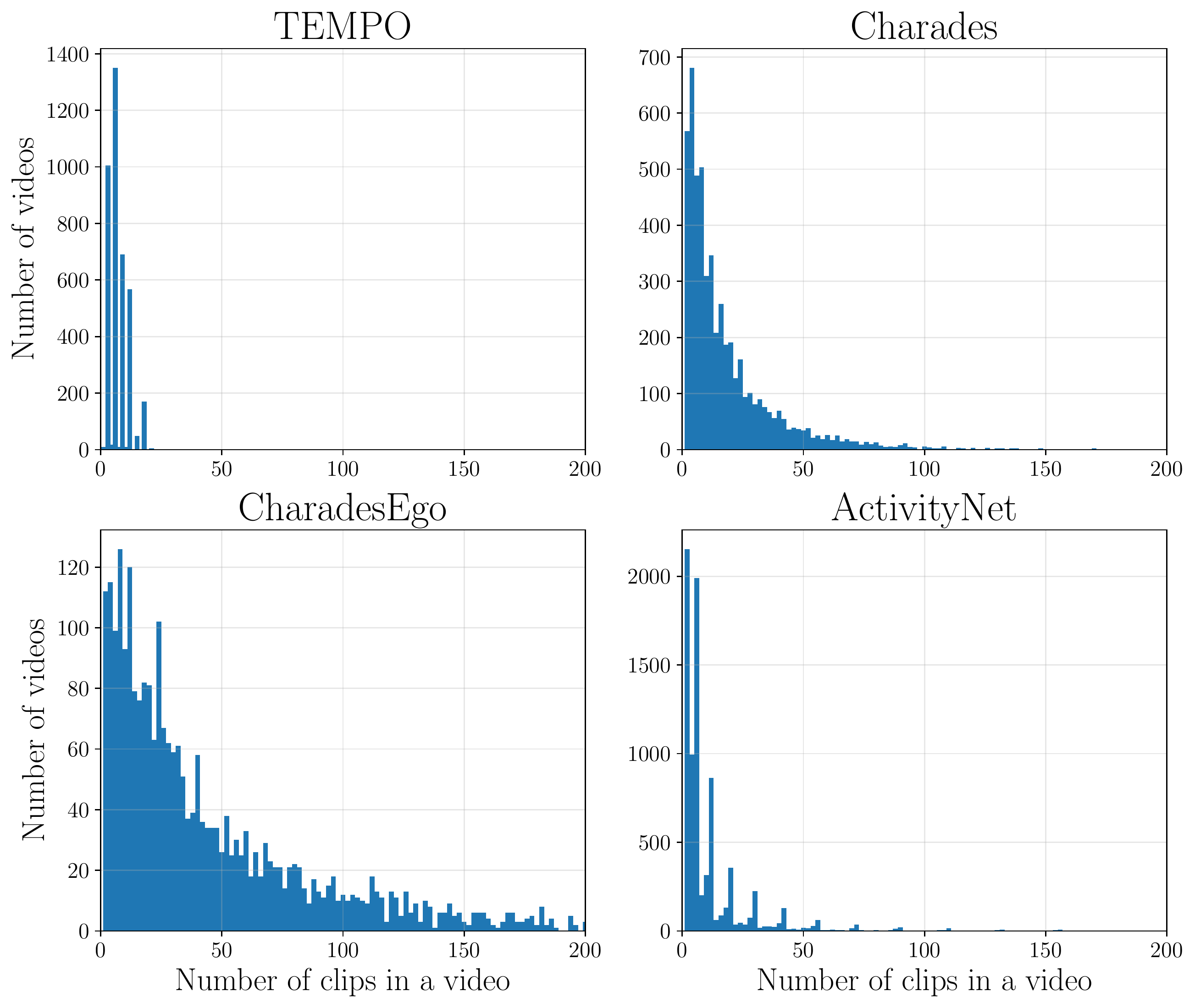}
    \caption{Number of clips in a video. The number of clips per video is lower
    in TEMPO and ActivityNet as compared to Charades and Charades-Ego.}
    \label{fig:dist_num_clips}
\end{figure}

\paragraph{Distribution of number of clips in a video.}
A single video with $10$ non-overlapping individual event clips can make upto
$^{10}C_{2} {=} 45$ stitched clips. We plot the number of clips per video
against the number of videos in a given dataset in \cref{fig:dist_num_clips}. A
single video with ${>} 30$ stitched clips is rare in TEMPO and ActivityNet while
much more frequent in Charades and Charades-Ego. Overall, the number of clips
per video is lower in TEMPO and ActivityNet as compared to Charades and
Charades-Ego.

\begin{figure}[!ht]
\captionsetup{skip=2mm,font=small}
\centering

\begin{subfigure}[b]{\columnwidth}
   \includegraphics[width=1\linewidth]{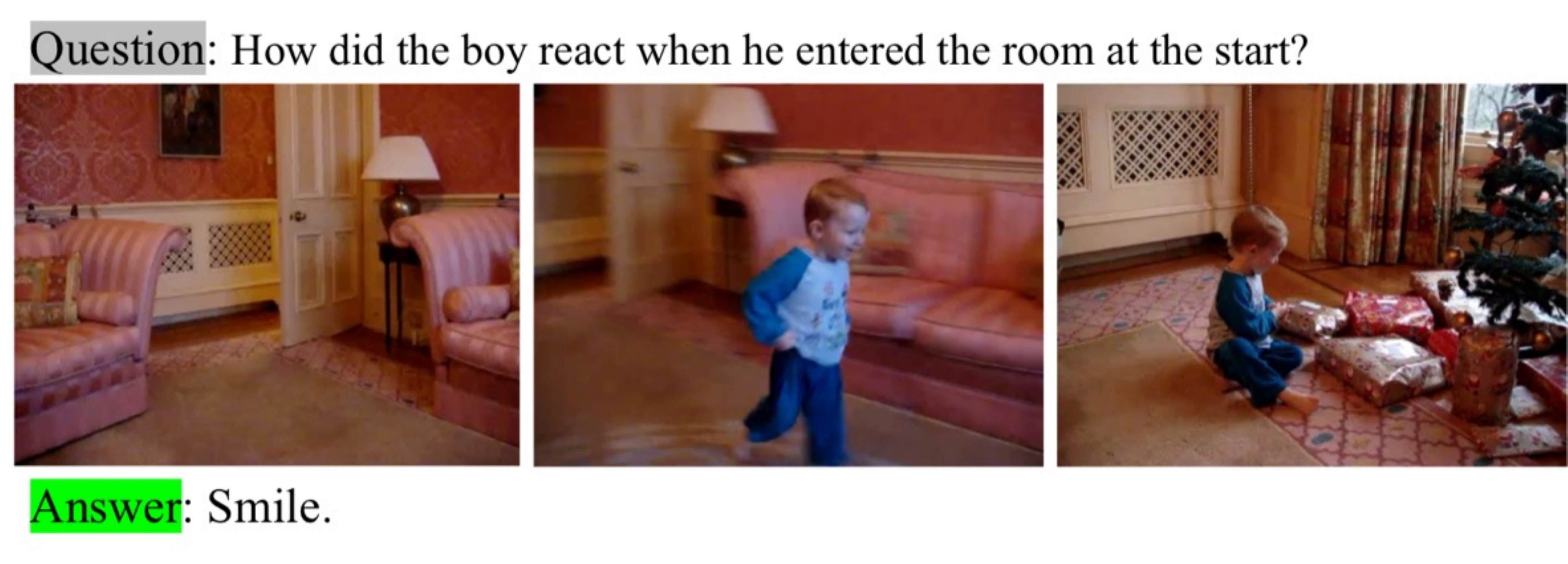}
\end{subfigure}

\begin{subfigure}[b]{\columnwidth}
\captionsetup{skip=0mm}
   \includegraphics[width=1\linewidth]{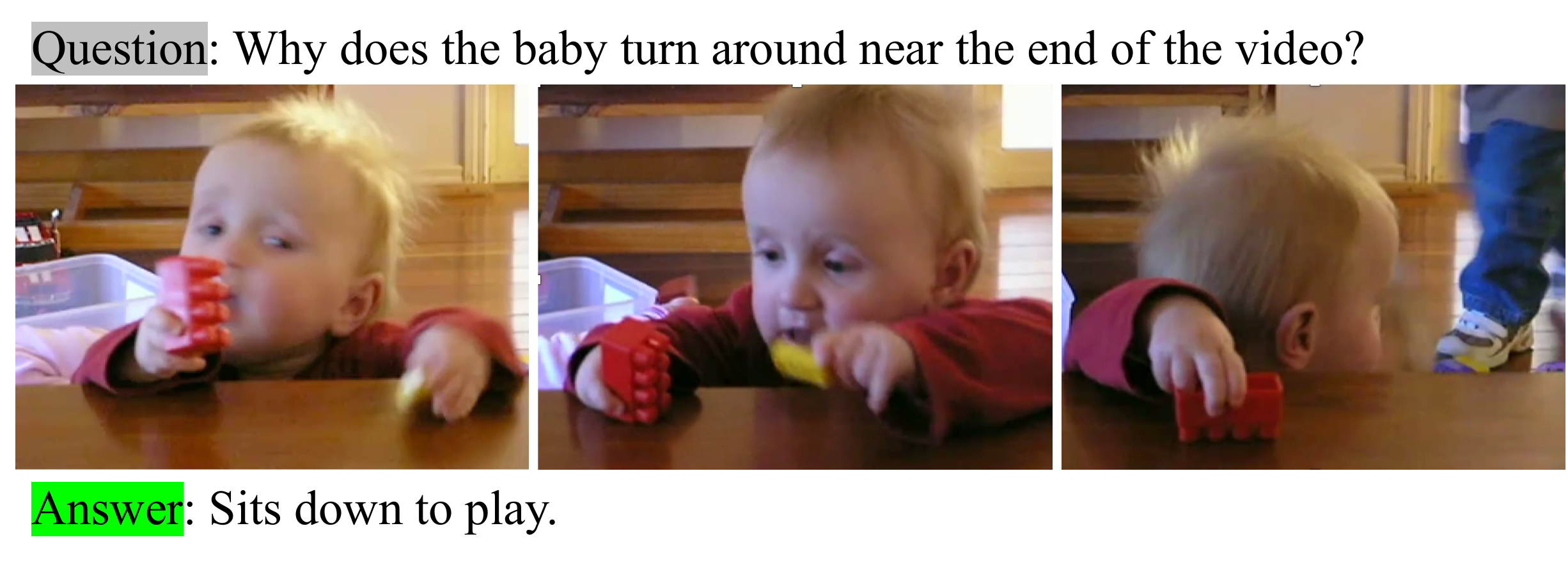}
   \caption{Next-QA: Video question answering}
   \label{fig:nextqa}
\end{subfigure}

\begin{subfigure}[b]{\columnwidth}
   \includegraphics[width=\linewidth]{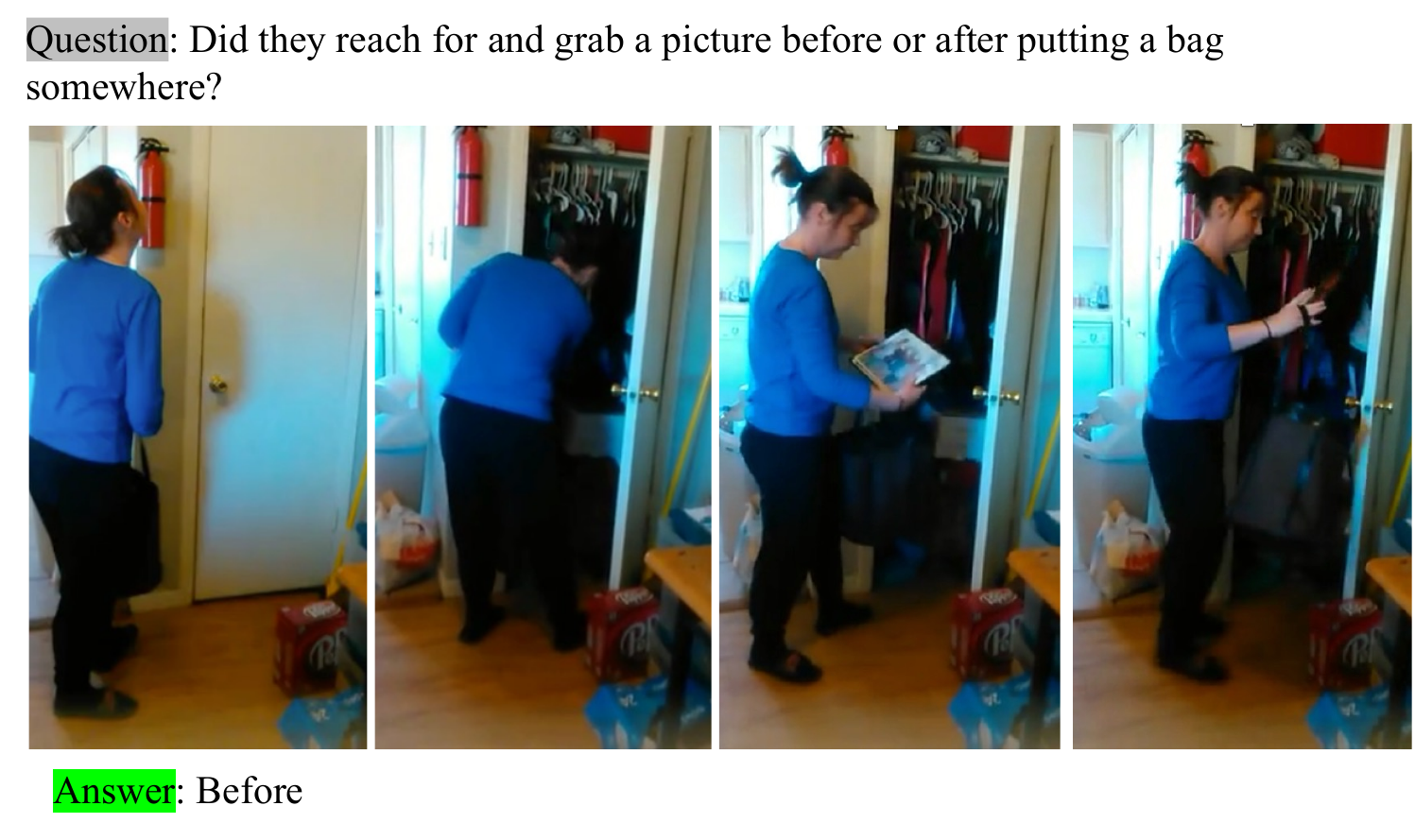}
\end{subfigure}
\begin{subfigure}[b]{\columnwidth}
\captionsetup{skip=-1mm}
   \includegraphics[width=1\linewidth]{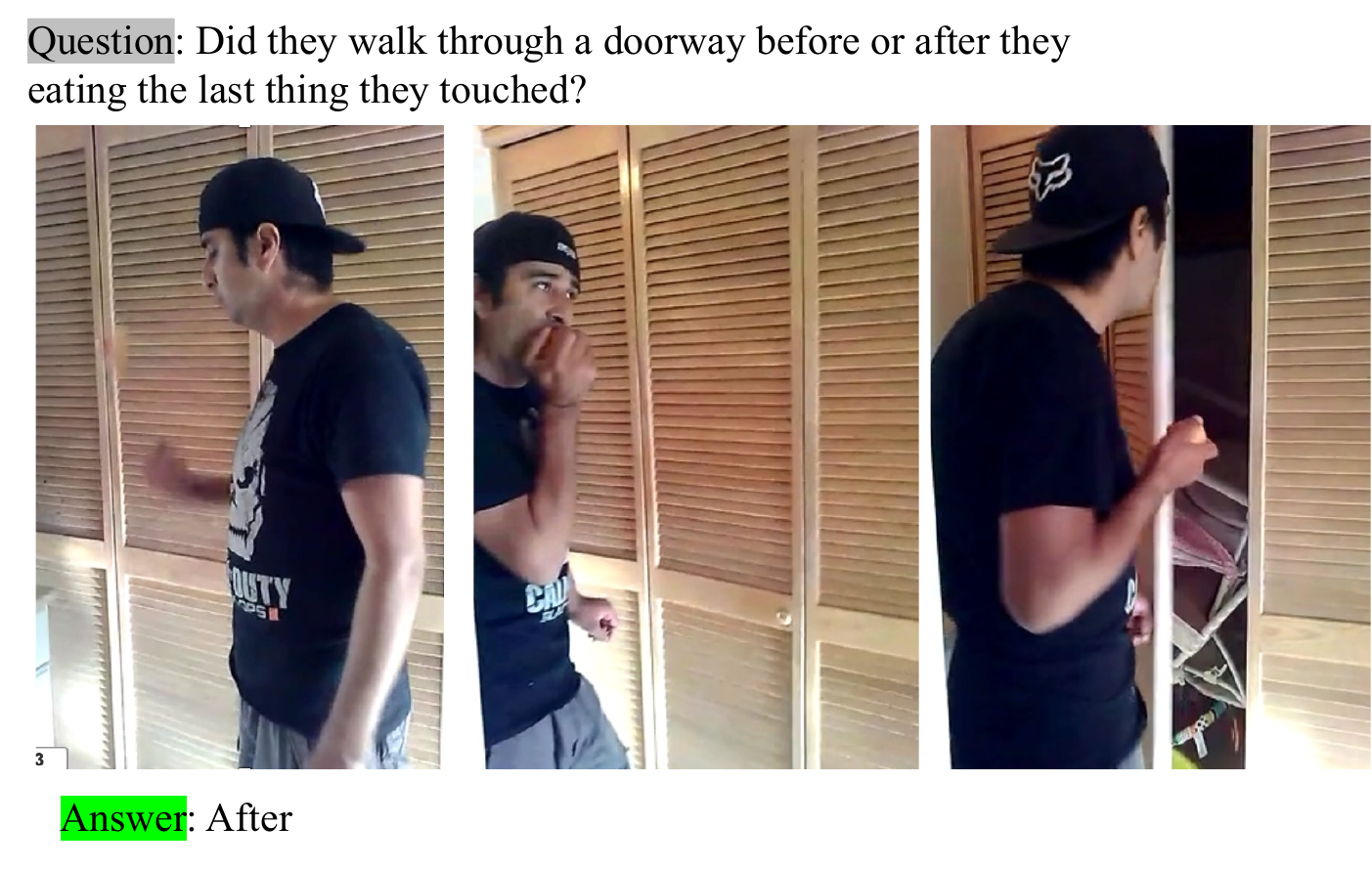}
   \caption{AGQA: Video question answering}
   \label{fig:agqa}
\end{subfigure}

\begin{subfigure}[b]{\columnwidth}
\captionsetup{skip=1mm}
   \includegraphics[width=1\linewidth]{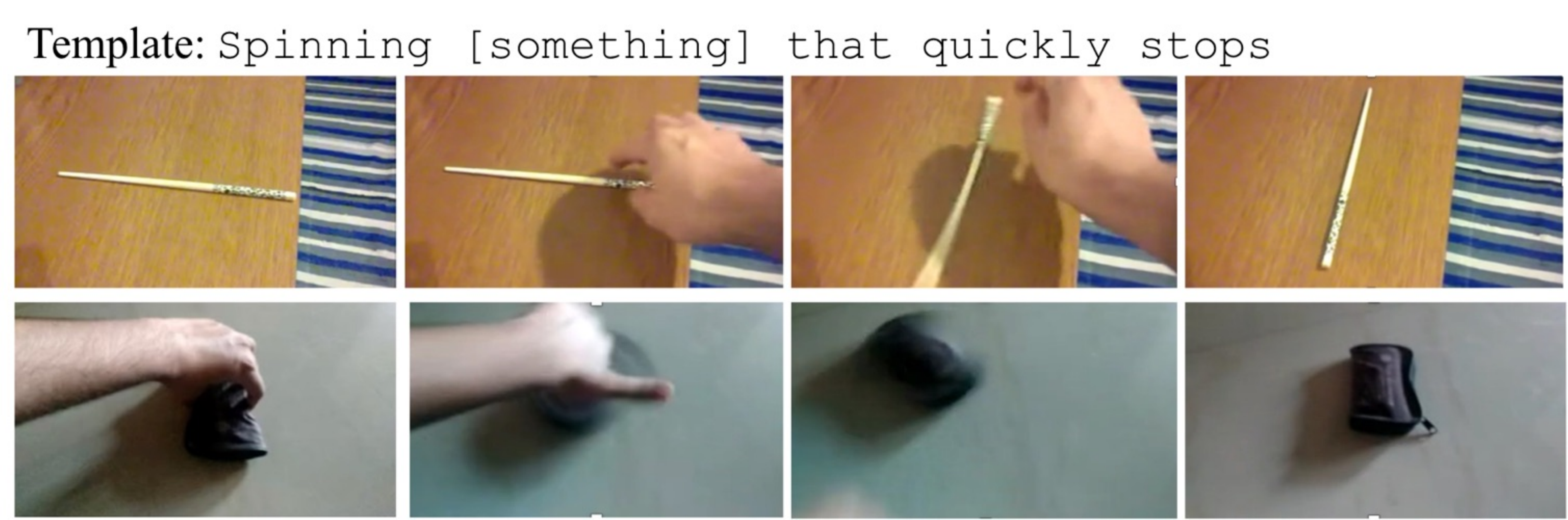}
   \caption{Something-Something: Template-based video retrieval}
   \label{fig:ssv2}
\end{subfigure}

\caption{Examples from datasets used for downstream evaluation. These tasks
   demand time awareness since it is often not possible to infer the action from
   a single frame.
}
\label{fig:downstream-datasets}
\end{figure}

\paragraph{Downstream datasets.}
In \cref{fig:downstream-datasets},  we also show some examples from some
downstream datasets (tasks) that need higher time awareness since they typically
involve multiple temporally linked events (\eg, \texttt{walk} and \texttt{eat}
in \cref{fig:downstream-datasets}(b)). On these datasets, we perform zero-shot
evaluation of temporally adapted models in \cref{sec:downstream} of the main paper.

\section{Experiments}
\label{sec:supp-expts}

\paragraph{Analyzing more pretrained models.}
\label{para:more-pretrained-models}
We present preliminary experimental results for other pretrained models on the
Charades dataset in \cref{tab:more-pretrained-models}. The other models perform
(slightly) better than random, but are not as promising as VideoCLIP. We observe
similar trends in performance on the TEMPO dataset. We hypothesize that
VideoCLIP’s larger temporal receptive field and contrastive pre-training
objective similar to TACT helps it achieve superior performance. This merits
further investigation into how various factors (as tabulated in
\cref{tab:more-pretrained-models}) influence temporal adaptation.
\input{tables/rebuttal-table-1}

\paragraph{Spatial~\vs~temporal understanding.}
\label{para:spatial-vs-temporal}
An interesting facet of TACT is $\alphasame$ which controls how well a model
adapts to temporal tasks. We highlight this on the TEMPO dataset in
\cref{tab:alpha-same-downstream}, where, $\alphasame {=} 0$ results in $\Atime$
$\sim$50\% while $\alphasame {=} 1$ improves performance. Further investigation
on downstream tasks shows that adaptation with $\alphasame{=}1$ does not perform
well on MSR-VTT (a non-temporal benchmark) but shows consistent improvements on
AGQA (a temporal benchmark).and the trade-off between spatial- and
temporal-understanding. This hints at $\alphasame$ controlling the trade-off
between spatial and temporal understanding.
\input{tables/6_downstream-alphasame}

\paragraph{What makes temporal adaptation difficult?}
\label{para:temporal-adaptation-hardness}
To recall, we define $\Deltatime$ as the time-distance (in seconds) between the
midpoints of the two clips in a stitched pair. We hypothesize that  $\Deltatime$
is inversely related to the difficulty of temporal adaptation, \ie, the larger
$\Deltatime$, the easier it is to distinguish between two stitched clips that
have opposite time order. For example, consider ActivityNet examples in
\cref{fig:examples}(b) where the visual context changes significantly making
inference of the time order of events relatively easier.

\begin{figure*}[ht]
\captionsetup{skip=-3mm,font=small}
\centering
\begin{subfigure}{0.8\textwidth}
  \centering
  \includegraphics[width=\linewidth]{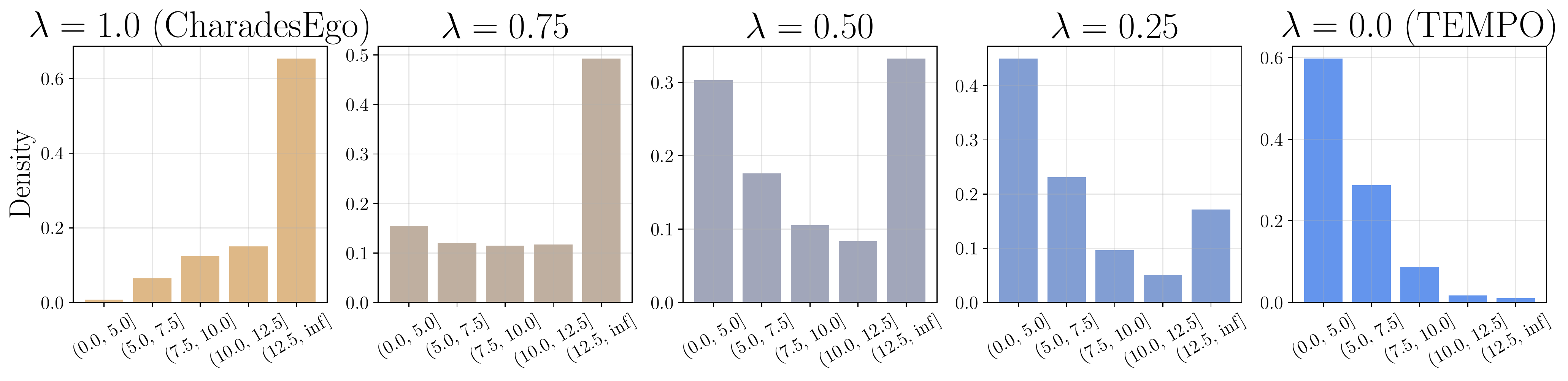}
  \label{fig:dist-shift-example}
\end{subfigure}%
\begin{subfigure}{.2\textwidth}
  \centering
  \includegraphics[width=\linewidth]{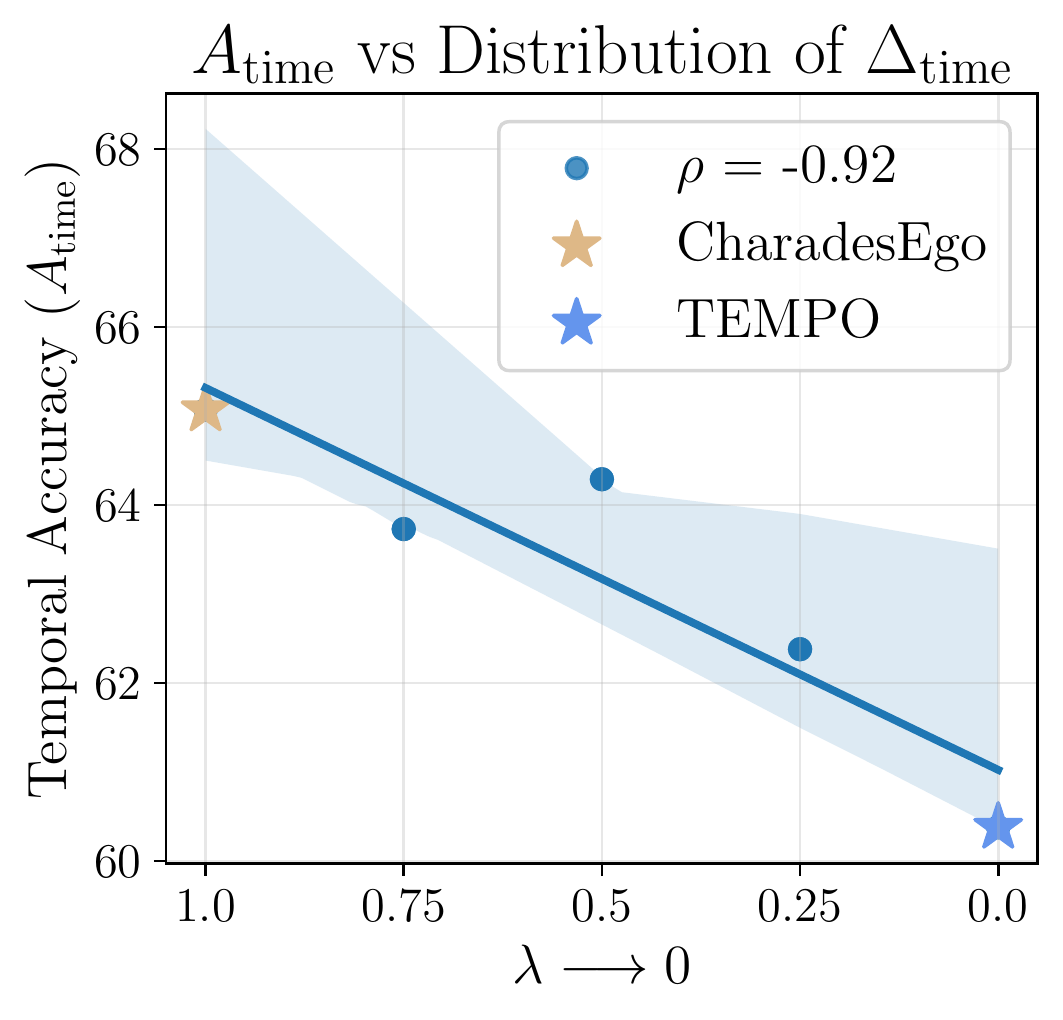}
  \label{fig:dist-shift-result}
\end{subfigure}
\caption{Impact of changing distribution of $\Deltatime$, the time gap between two stitched clips.
\textbf{Left:}
We vary the distribution of $\Deltatime$ for Charades-Ego and make it similar to
that of TEMPO as $\lambda \rightarrow 0$. Thus, crudely, as $\lambda$ decreases,
so does $\Deltatime$. \textbf{Right:} $\Atime$ on Charades-Ego where the time
difference between sampled clips is according to the distributions on the left.
We observe that the accuracy deteriorates as the time-distance between a pair of
clips decreases indicating a strong correlation between the distribution of
$\Deltatime$ and difficulty of temporal adaptation.}
\label{fig:dist-shift}
\end{figure*}

We further test our hypothesis by sampling individual clips from the
Charades-Ego dataset to match the $\Deltatime$ distribution of TEMPO.
Concretely, assuming $\Deltatime$ for both these datasets follows a multinomial
distribution, we construct a new distribution using a convex combination of the
individual distributions where the mixing parameter $\lambda \in [0, 1]$
controls the extent to which we modify the distribution from TEMPO
($\lambda{=}0$) to Charades-Ego ($\lambda{=}1$). The resulting distributions are
presented in \cref{fig:dist-shift} (left). With $\lambda{=}1$, we sample from
the original Charages-Ego distribution and gradually move towards TEMPO as
$\lambda \rightarrow 0$. 

We then sample stitched clips according to this new distribution and
post-pretrain temporal adaptation for varying values of $\lambda$. Note that for
this experiment, we keep fixed $N_{\mathcal{C}} {=} 10,000$ for each $\lambda$.
From \cref{fig:dist-shift} (right), we indeed find that as we move towards a
more TEMPO-like distribution (shorter $\Deltatime$), temporal accuracy
deteriorates. The best fit also confirms that the distribution of $\Deltatime$
is strongly correlated ($\rho = -0.92$) with the difficulty of inferring
time-order consistency.

\section{Qualitative Analysis}
\label{sec:qualitative-analysis}

\begin{figure}[ht!]
    \centering
    \includegraphics[width=\linewidth]{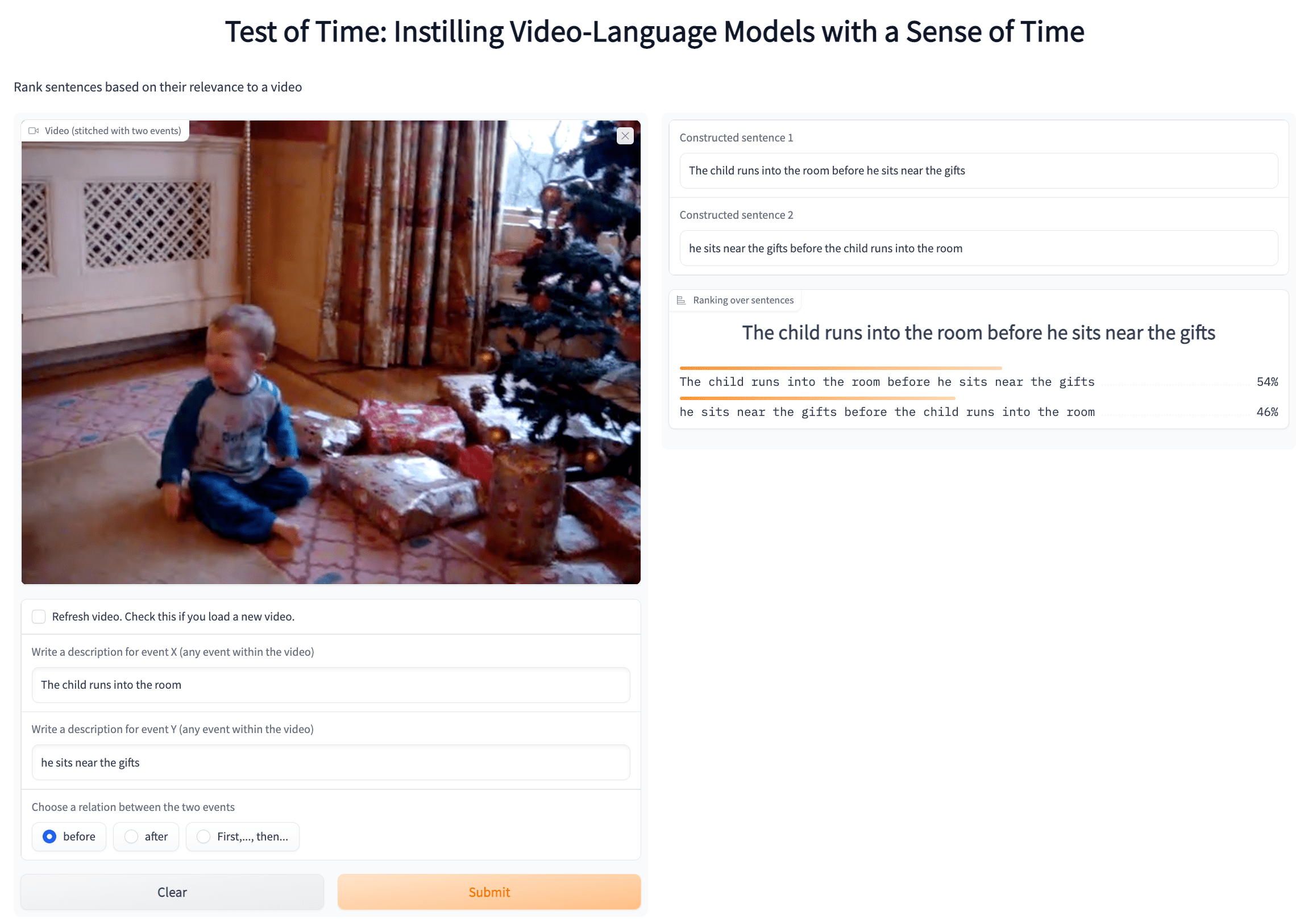}
    \caption{Interface of our demo for qualitative analysis. The user uploads a
      video and is asked to describe two events in the video. These event
      descriptions are then connected via one of the three temporal relations
      shown at the bottom left. We construct one sentence that is consistent
      with the time order of events in the video and another that is not. The
      output on the right shows the ranking of the constructed sentences in
      terms of cosine similarity with the video representation. Higher score for
      correct matching indicated by a longer orange bar. Best viewed zoomed in.}
    \label{fig:demo}
\end{figure}

To get an intuitive sense of whether a TACT model understands  time order of
events, we perform a qualitative analysis on the model trained on TEMPO. Our
demo interface looks like the one shown in \cref{fig:demo}. First, a user
uploads a video and adds text descriptions for two events within the video.
These descriptions are then connected via a temporal relation such as
\texttt{before} or \texttt{after}. We also experiment with a new temporal
connector \texttt{First, ..., then, ....} to check if our model generalizes
beyond before/after.

\begin{figure}
\captionsetup{skip=2mm,font=small}
\centering

\begin{subfigure}{\columnwidth}
  \centering
  \includegraphics[width=\linewidth]{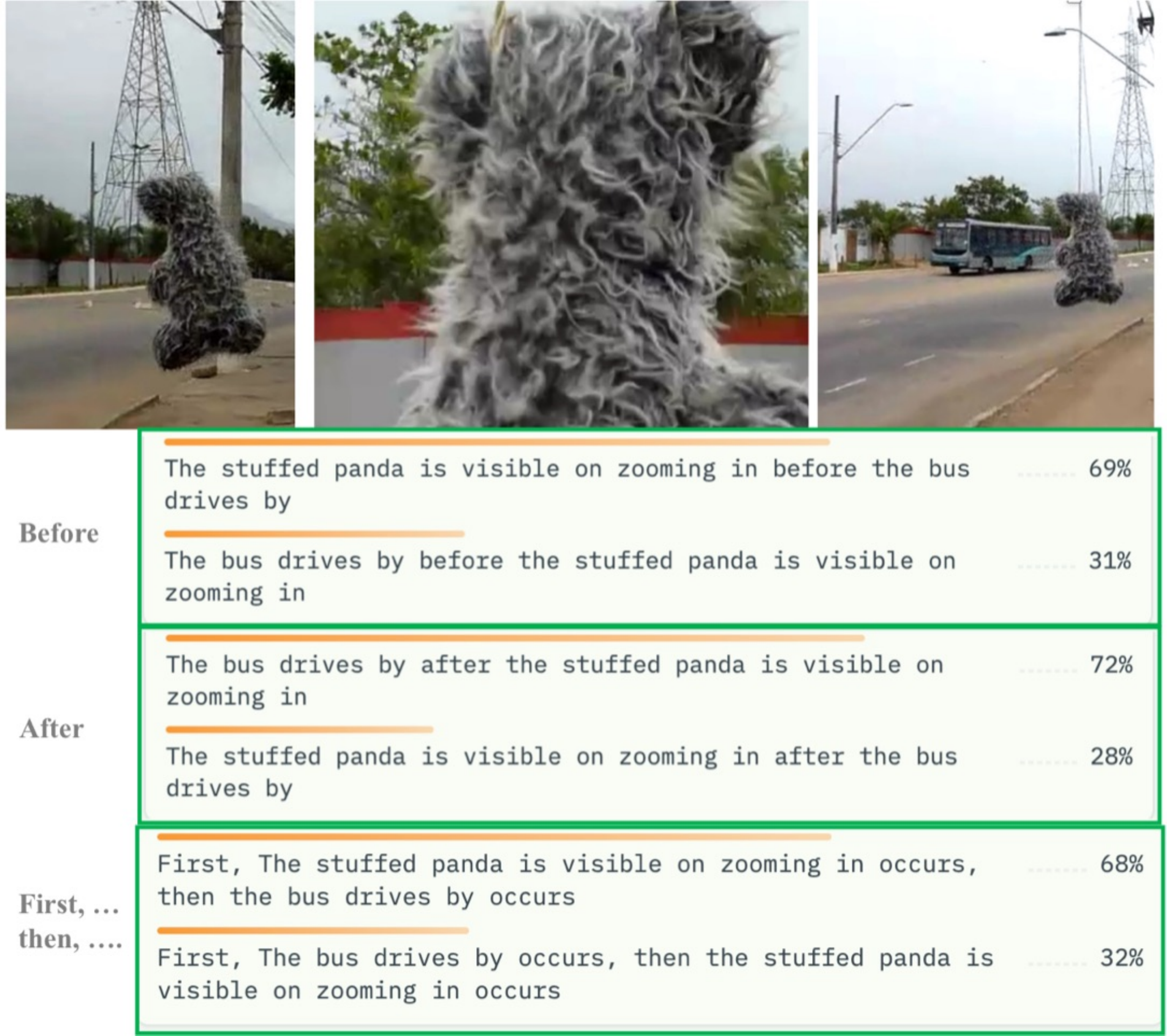}
\end{subfigure}%
\vspace{2mm}

\begin{subfigure}{\columnwidth}
  \centering
  \includegraphics[width=\linewidth]{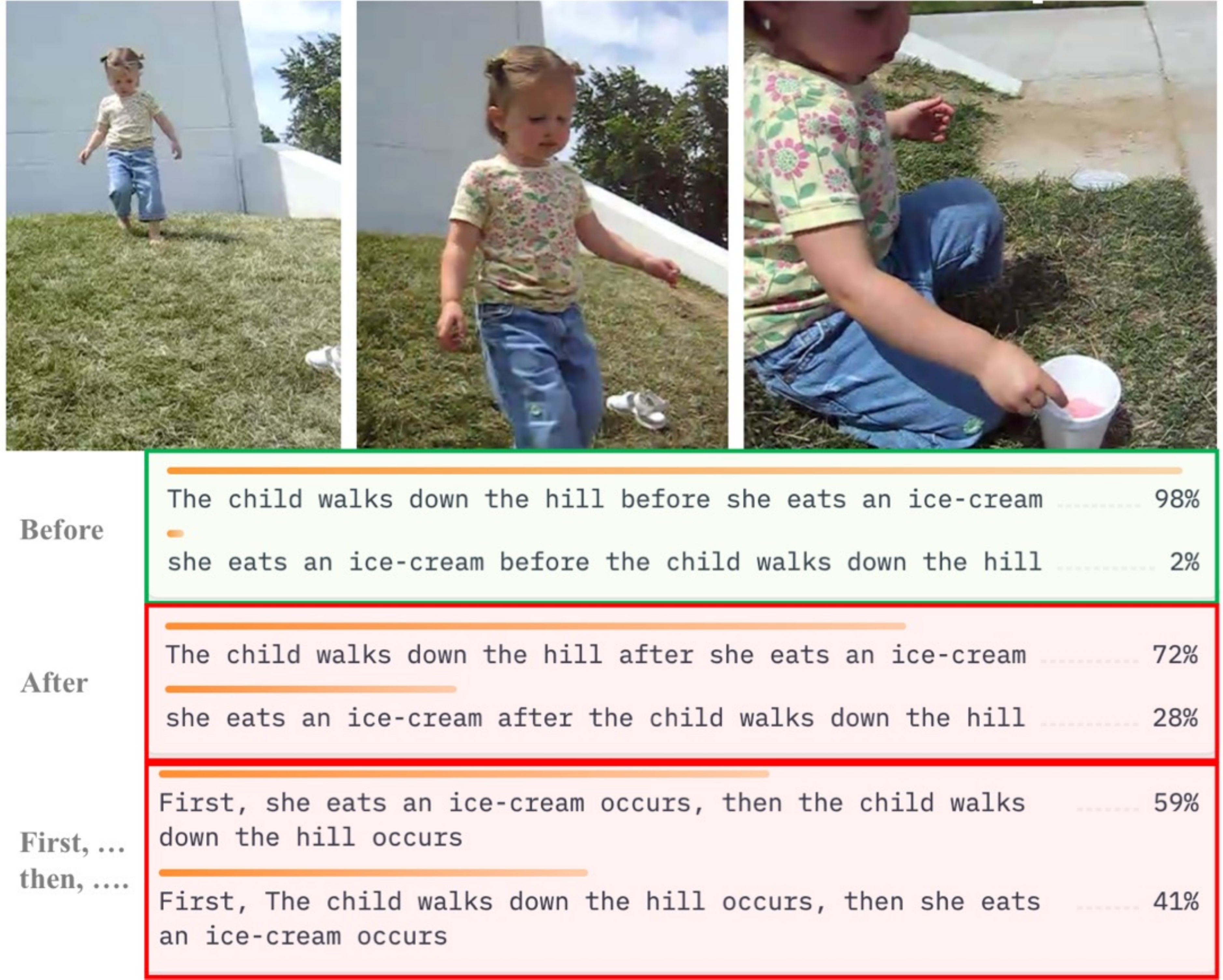}
\end{subfigure}%

\caption{Qualitative examples from TEMPO validation set. We evaluate similarity
of a given video with sentences with different temporal order with the usual
temporal connectors (before/after). \colorbox{green!15}{Green} bordered boxes
indicate correct predictions (consistent time order between video and language)
while \colorbox{red!15}{red} denote mispredictions. For some examples, \eg, in
the bottom example, the model gets predictions incorrect particularly for
relations other than \texttt{before}. Furthermore, we also try a new temporal
connector \texttt{First, ..., then, ...} and observe that the model
qualitatively generalizes to that as well.}
\label{fig:qual-tempo-multiple}
\end{figure}

\begin{figure}
\centering
\captionsetup{skip=2mm,font=small}

\begin{subfigure}{\columnwidth}
  \centering
  \includegraphics[width=\linewidth]{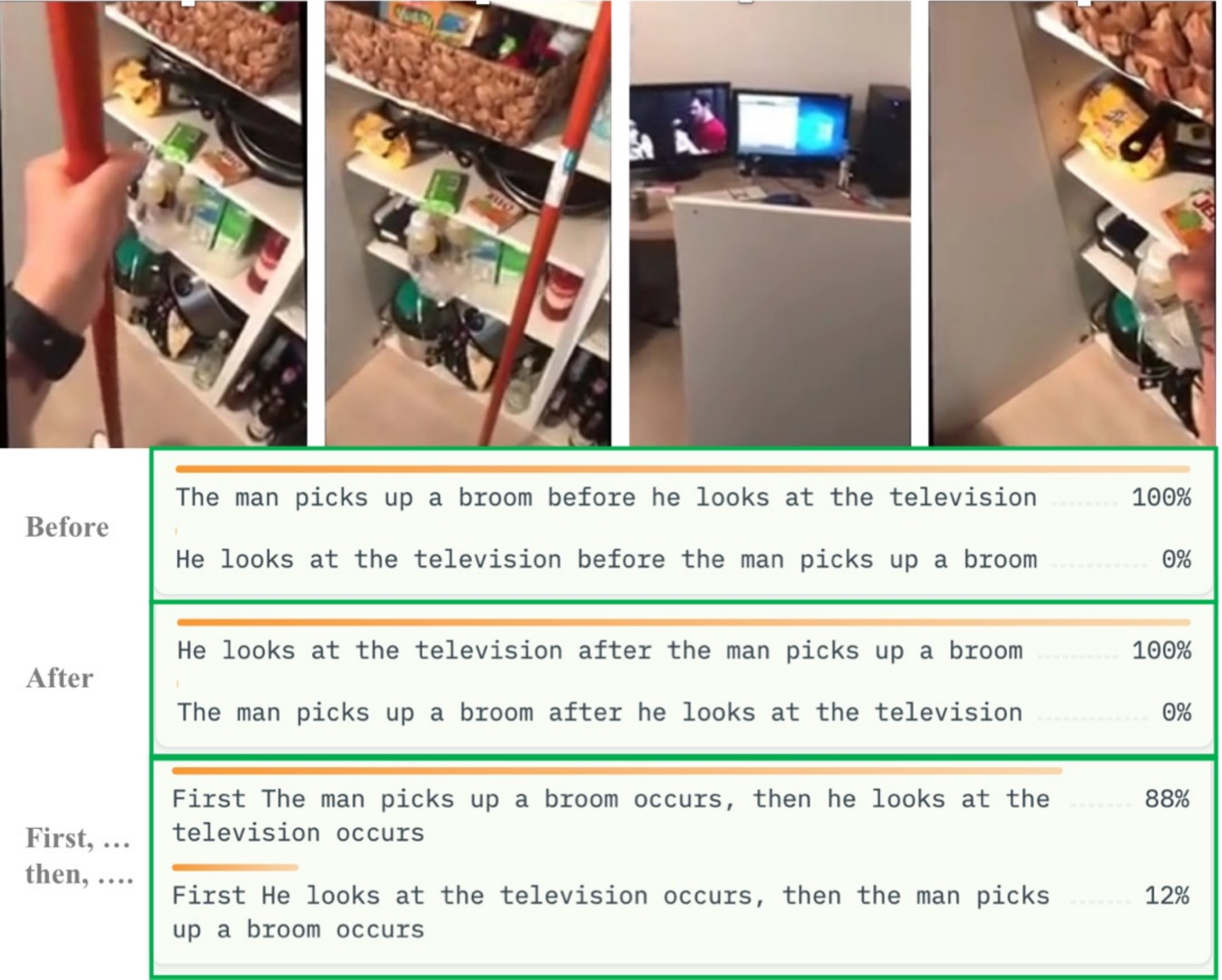}
  \caption{Example from Charades-Ego}
  \label{fig:qual-charadesego}
\end{subfigure}%
\vspace{2mm}

\begin{subfigure}{\columnwidth}
  \centering
  \includegraphics[width=\linewidth]{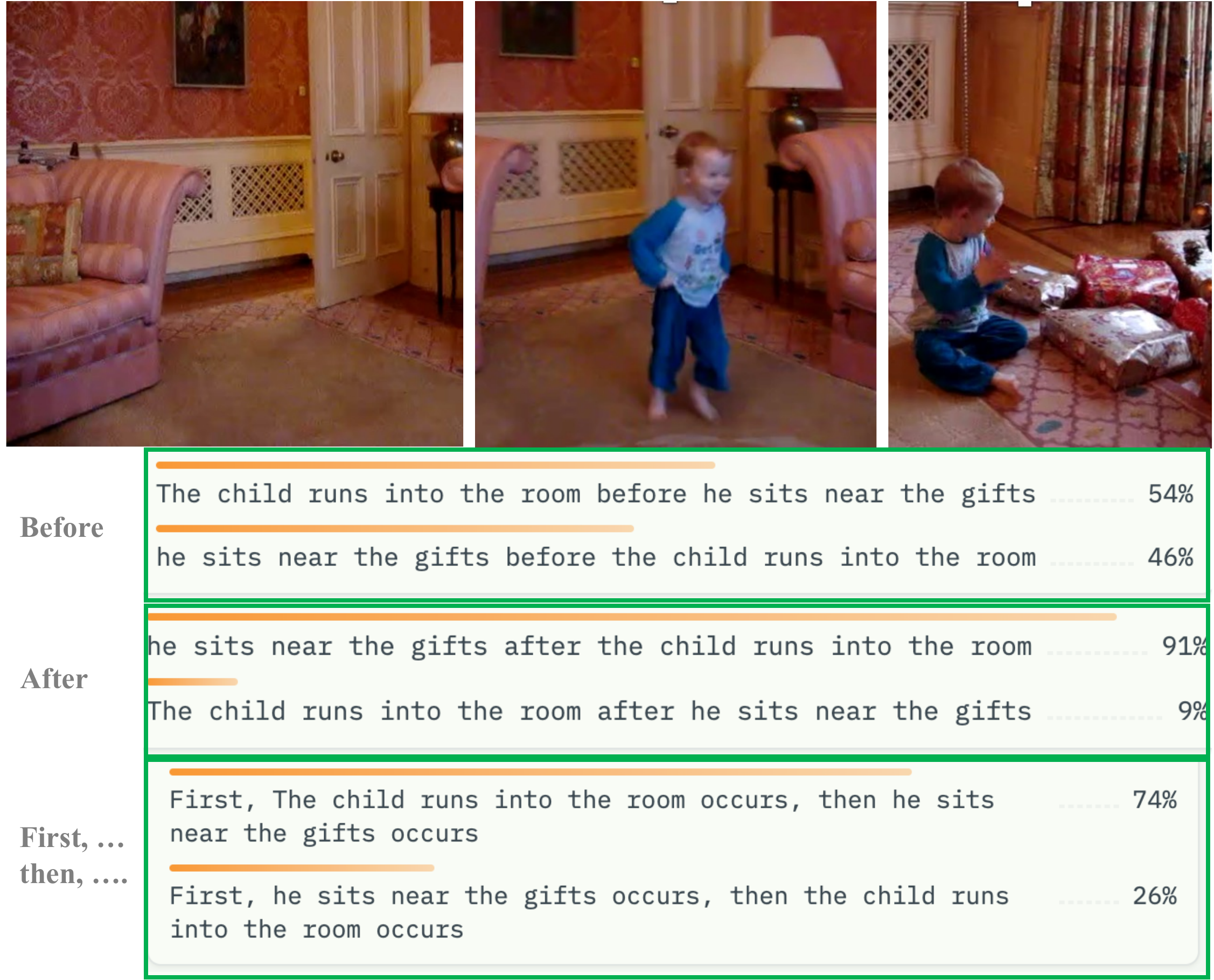}
  \caption{
    Example from Next-QA
  }
  \label{fig:qual-nextqa}
\end{subfigure}%

\caption{Qualitative results on samples \textit{\textbf{not}} from TEMPO. We
   see that despite not having seen these examples, the model still connects the
   time order across video and language correctly.}
\label{fig:qual-others}
\end{figure}

First, we consider samples from the TEMPO validation set and show their results
in \cref{fig:qual-tempo-multiple}. Notably, for some examples, it connects time
order for \texttt{before} relations but not the other two. We suspect this is
because a majority ($\sim60\%$) of the TEMPO dataset has descriptions involving
\texttt{before}. Note that TEMPO already comes with temporal captions of which
we pick subset of before/after relations. Second, we also consider samples from
other datasets which the model has never seen. To our surprise, albeit
qualitatively, the model does generalize well to such examples as shown in
\cref{fig:qual-others}. 

These results reinforce the promise of our method and also raise the possibility
of extending this work to consider more general temporal relations. Having said
that, we reiterate that these are qualitative examples and should be treated as
such.

%% file: tables/rebuttal-table-1.tex
\begin{table}[b]
\centering
\tabcolsep=0.10cm
\captionsetup{skip=2mm}
\resizebox{\columnwidth}{!}{%
\begin{tabular}{lccccr}
\toprule
\textbf{Model} & \textbf{\begin{tabular}[c]{@{}c@{}}Temporal\\ receptive field\end{tabular}} & \textbf{\begin{tabular}[c]{@{}c@{}}Pre-training\\ strategy\end{tabular}} & \textbf{Visual backbone} & \textbf{\begin{tabular}[c]{@{}c@{}}Encoder\\ \end{tabular}} & \multicolumn{1}{l}{$\mathbf\Atime$} \\
\midrule
Frozen~\cite{Bain21-frozen} & 4 & Contrastive & TimeSformer & Multimodal & 53.0 \\
VindLU~\cite{cheng2022vindlu} & 8 & Autoencoding & ViT{+}Temp. attn. & Multimodal & 54.1 \\
CLIP4Clip~\cite{Luo2021-CLIP4Clip} & 12 & Contrastive & ViT{+}Temp. attn. & Two-tower & 57.5 \\
VideoCLIP~\cite{xu-etal-2021-videoclip} & \textbf{32} & Contrastive & BERT on S3D & Two-tower & \textbf{77.0} \\
\bottomrule
\end{tabular}%
}
\caption{ Adaptation results for more pre-trained models on Charades. Models with smaller temporal receptive field perform worse in comparison to VideoCLIP. The temporal receptive field is reported in terms of the number of input frames. Systematically understanding the influence of various factors on making models time-aware by post-pretraining makes for interesting future work. }
\label{tab:more-pretrained-models}
\end{table}

%% file: tables/6_downstream-alphasame.tex
\begin{table}[t!]
\captionsetup{skip=2mm,font=small}
\centering
\resizebox{\linewidth}{!}{%
\begin{tabular}{ccc r c rrr}
\toprule
\multicolumn{3}{c}{\textbf{Hyperparameters}} & \multicolumn{1}{c}{\textbf{\ \ Adaptation}}& & \multicolumn{3}{c}{\textbf{Downstream}} \\
\midrule
$\alphasame$ & $\alphacross$ & $\beta$ & \multicolumn{1}{c}{TEMPO} & & \multicolumn{2}{c}{MSR-VTT} & \multicolumn{1}{c}{AGQA} \\
\multicolumn{1}{r}{} & \multicolumn{1}{r}{} & \multicolumn{1}{r}{} & $\Atime \uparrow$ & & $R@1 \uparrow$ & MedR $\downarrow$ & Accuracy$\uparrow$ \\
\midrule
0 & 0 & 0 & 49.4 & & \cellcolor{gray!15}15.0 & 20.0 & 50.5 \\
0 & 0 & 1 & 49.5 & & \cellcolor{gray!15}14.2 & 20.0 & 49.9 \\
0 & 1 & 0 & 49.3 & & \cellcolor{gray!15}14.4 & 19.0 & 50.2 \\
0 & 1 & 1 & 49.5 & & \cellcolor{gray!15}15.1 & 19.0 & 50.2 \\
\arrayrulecolor{lightgray}\midrule
1 & 0 & 0 & \cellcolor{gray!15}60.6 & & 11.7 & 27.0 & \cellcolor{gray!15}56.6 \\
1 & 0 & 1 & \cellcolor{gray!15}62.9 & & 9.4 & 36.0 & \cellcolor{gray!15}58.3 \\
1 & 1 & 0 & \cellcolor{gray!15}59.7 & & 9.1 & 37.0 & \cellcolor{gray!15}56.9 \\
1 & 1 & 1 & \cellcolor{gray!15}62.5 & & 12.8 & 27.0 & \cellcolor{gray!15}57.1 \\
\arrayrulecolor{black}\bottomrule
\end{tabular}%
}
\caption{Impact of $\alphasame$ on spatial- \vs temporal understanding. \colorbox{gray!15}{Gray} denotes better performance 
for
$\alphasame{=}0$ or $1$. While $\alphasame{=}1$ drives temporal understanding, it comes at a cost of retrieval performance on MSR-VTT~\cite{xu2016-msr-vtt}. 
This hints at $\alphasame$ controlling the trade-off between spatial- and temporal-understanding.
}
\label{tab:alpha-same-downstream}
\end{table}